\documentclass[moor]{informs1}              

\usepackage{algorithm, algorithmic}
\usepackage{caption}
\usepackage{subcaption}
{\theoremstyle{THkey}\newtheorem{mytheorem}{XXXXX}}

\usepackage[sort]{natbib}
 \NatBibNumeric
 \def\bibfont{\small}%
 \bibpunct[, ]{[}{]}{,}{n}{}{,}%

\usepackage[colorlinks=true,breaklinks=true,bookmarks=true,urlcolor=blue,
     citecolor=blue,linkcolor=blue,bookmarksopen=false,draft=false]{hyperref}

\def\EMAIL#1{\href{mailto:#1}{#1}}

\TheoremsNumberedThrough     

\EquationsNumberedThrough    

\begin{document}


\RUNAUTHOR{Russo and Van Roy}

\RUNTITLE{Learning to Optimize Via Posterior Sampling}

\TITLE{Learning to Optimize Via Posterior Sampling}

\ARTICLEAUTHORS{%
\AUTHOR{Daniel Russo}
\AFF{Department of Management Science and Engineering, Stanford University, Stanford, California, 94305, \EMAIL{djrusso@stanford.edu}}
\AUTHOR{Benjamin Van Roy}
\AFF{Departments of Management Science and Engineering and Electrical Engineering, Stanford University, Stanford, California, 94305, \EMAIL{bvr@stanford.edu}}
} 

\ABSTRACT{%
This paper considers the use of a simple posterior sampling algorithm to balance between exploration and exploitation when learning to optimize actions such as in multi-armed bandit problems. The algorithm, also known as {\it Thompson Sampling} and as {\it probability matching}, offers significant advantages over the popular upper confidence bound (UCB) approach, and can be applied to problems with finite or infinite action spaces and complicated relationships among action rewards.  We make two theoretical contributions. The first establishes a connection between posterior sampling and UCB algorithms. This result lets us convert regret bounds developed for UCB algorithms into Bayesian regret bounds for posterior sampling. Our second theoretical contribution is a Bayesian regret bound for posterior sampling that applies broadly and can be specialized to many model classes. This bound depends on a new notion we refer to as the {\it eluder dimension},
which measures the degree of dependence among action rewards.  Compared to UCB algorithm Bayesian regret bounds for specific model classes,
our general bound matches the best available for linear models and is stronger than the best available for generalized linear models.
Further, our analysis provides insight into performance advantages of posterior sampling, which are highlighted through simulation results that demonstrate performance surpassing recently proposed UCB algorithms.
}%


\MSCCLASS{Primary: 93E35; secondary: 62L05}
\ORMSCLASS{Primary: decision analysis: sequential;} 
\HISTORY{Received February 26, 2013; revised November 21, 2013}

\maketitle

%


\section{Introduction.}
We consider an optimization problem faced by an agent who is uncertain about how his actions influence performance.   
The agent selects actions sequentially, and upon each action observes a reward. A 
{\it reward function} governs the mean reward of each action.  The agent 
represents his initial beliefs through a prior distribution over reward functions.  As rewards
are observed the agent learns about the reward function, and this allows him to improve his behavior.  
Good performance requires adaptively sampling actions in a way that strikes an effective balance between 
exploring poorly understood actions  and  exploiting previously 
acquired knowledge to attain high rewards.  In this paper, we study a simple algorithm for 
selecting actions and provide finite time performance guarantees that apply
across a broad class of models.

The problem we study has attracted a great deal of recent interest and is often referred to as the multi-armed 
bandit (MAB) problem with dependent arms.  We refer to the problem as one of \textit{learning to optimize} to
emphasize its divergence from the classical MAB literature. In the typical MAB framework, 
there are a finite number of actions that are modeled independently; 
sampling one action provides no information about the rewards that can be gained through selecting other actions.
In contrast, we allow for infinite action spaces and for general forms
of model uncertainty, captured by a prior distribution over a set of possible reward functions.
Recent papers have addressed this problem in cases where the relationship among
action rewards takes a known parametric form. For example, \citet{dani2008stochastic, abbasi2011improved,  rusmevichientong2010linearly}
study the case where actions are described by a finite number of features
and the reward function is linear in these features. Other authors have studied cases where
the reward function is Lipschitz continuous \cite{kleinberg2008multi, bubeck2011xarmed, valko2013stochastic}, sampled
from a Gaussian process \cite{srinivas2012information}, or takes the form of a generalized
\cite{filippi2010parametric} or sparse \cite{abbasi2012online} linear model.

Each paper cited above studies an upper confidence bound (UCB) 
algorithm.  Such an algorithm forms an optimistic estimate of the mean-reward
value for each action, taking it to be the highest statistically plausible 
value. It then selects an action that maximizes among these optimistic
estimates.  Optimism encourages selection of poorly-understood 
actions, which leads to informative observations.  As data accumulates, optimistic estimates 
are adapted, and this process of exploration and learning converges toward optimal behavior.

We study an alternative algorithm that we refer to as \textit{posterior sampling}. It is also
also known as \textit{Thompson sampling} and as \textit{probability matching}.
The algorithm randomly selects an action according to the probability
it is optimal. Although posterior sampling was first proposed
almost eighty years ago, it has until recently received little
attention in the literature on multi-armed bandits. While its asymptotic convergence has been
established in some generality \cite{may2012optimistic},
not much else is known about its theoretical properties in the case of dependent
arms, or even in the case of independent arms with general prior distributions.
Our work provides some of the first theoretical guarantees.

Our interest in posterior sampling is motivated by several potential advantages
over UCB algorithms, which we highlight in Section \ref{subsec:advantages}. While particular UCB algorithms can be extremely
effective, performance and computational tractability depends critically on the confidence sets used by the algorithm. 
For any given model, there is a great deal of design flexibility in choosing the structure of these sets. 
Because posterior sampling avoids the need for confidence bounds, its use greatly 
simplifies the design process and admits practical implementations in cases where UCB algorithms are computationally 
onerous. In addition, we show through simulations that posterior sampling outperforms various UCB algorithms that have 
been proposed in the literature. 

In this paper, we make two theoretical contributions. The first establishes a connection between posterior sampling and UCB algorithms.  
In particular, we show that while the regret of a UCB algorithm can be bounded in terms of the confidence bounds used by the algorithm, the Bayesian regret of posterior sampling can be bounded in an analogous way by {\it any} sequence of confidence bounds. In this sense, posterior sampling preserves many of the appealing theoretical properties of UCB algorithms without requiring explicit, designed, optimism. We show that, due to this connection, existing analysis available for specific UCB algorithms immediately translates to Bayesian regret bounds for posterior sampling.

Our second theoretical contribution is a Bayesian regret bound for posterior sampling 
that applies broadly and can be specialized to many specific model classes.  
Our bound depends on a new notion of dimension that measures the degree of
dependence among actions. We compare our notion of dimension to 
the Vapnik-Chervonenkis dimension and explain why that and other measures of 
dimension used in the supervised learning literature do not suffice when it 
comes to analyzing posterior sampling.

The remainder of this paper is organized as follows.  The next section
discusses related literature.  Section \ref{sec:formulation} then provides a formal problem statement.  We describe UCB
and posterior sampling algorithms in Section \ref{sec: algorithms}.  
We then establish in Section \ref{sec:implicit} a connection between them, which 
we apply in Section \ref{sec:UCB} to convert existing bounds for UCB algorithms to bounds for posterior sampling.
Section \ref{sec:dimension} develops a new notion of dimension and presents Bayesian regret bounds that depend on it. 
Section \ref{sec:simulation} presents simulation results.  A closing section makes concluding remarks.

\section{Related Literature.}\label{sec:literature}
One distinction of results presented in this paper is that they center around Bayesian regret as a measure of performance.  In the next subsection, 
we discuss this choice and how it relates to performance measures used in other work.  Following that, we review prior results and their relation to results of this paper.

\subsection{Measures of Performance.}

Several recent papers have established theoretical results on posterior sampling. 
One difference between this work and ours is that we focus on a different measure 
of performance. These papers all study the algorithm's regret, which measures its cumulative loss relative to an algorithm that always selects the 
optimal action, for some fixed reward function. To derive these bounds, each paper fixes an uninformative prior distribution with a convenient analytic structure,
and studies posterior sampling assuming this particular prior is used.  With one exception \cite{agrawal2013linear}, the focus is on the classical multiarmed bandit problem, where sampling
one action provides no information about others. 

Posterior sampling can be applied to a much broader class of problems, and one of its greatest strengths is its ability to incorporate prior knowledge in a flexible and coherent way. We therefore aim to develop results that accommodate the use of a wide range of models. Accordingly, most of our results allow for an {\it arbitrary} prior distribution over a particular class of mean reward functions. In order to derive meaningful results at this level of generality, we study the  algorithm's expected regret, where the expectation is taken with respect to the prior distribution over reward functions. This quantity is sometimes called the algorithm's {\it Bayesian regret}. We find this to be a practically relevant measure of performance and find this choice allows for more elegant analysis. Further, as we discuss in Section \ref{sec:formulation}, the Bayesian regret bounds we provide in some cases immediately yield regret bounds. 

In addition, studying Bayesian regret reveals deep connections between posterior sampling and the principle of {\it optimism in the face of uncertainty}, which we feel provides new conceptual insight into the algorithm's performance. Optimism in the face of uncertainty is a general principle and is not inherently tied to any measure of performance. Indeed,  algorithms based on this principle have been shown to be asymptotically efficient in terms of both regret \cite{lai1985asymptotically} and Bayesian regret \cite{lai1987adaptive},  to satisfy order optimal minimax regret bounds \cite{audibert2009minimax}, to satisfy order optimal bounds on regret and Bayesian regret when the reward function is linear \cite{rusmevichientong2010linearly}, and to satisfy strong bounds when the reward function is sampled from a Gaussian process prior \cite{srinivas2012information}. We take a very general view of optimistic algorithms, allowing upper confidence bounds to be constructed in an essentially arbitrary way based on the algorithm's observations and possibly the prior distribution over reward functions. 

\subsection{Related Results.}

Though it was first proposed in 1933, posterior sampling has until recently received
relatively little attention. Interest in the algorithm grew
after empirical studies \cite{chapelle2011empirical, scott2010modern} demonstrated 
performance exceeding state-of-the-art methods. An asymptotic convergence result was
established by \citet{may2012optimistic}, but finite time guarantees remain limited. 
The development of further performance bounds was raised as an open problem at the 
2012 Conference on Learning Theory \cite{liopen2012}.

Three recent papers 
\cite{kaufmann2012thompson,agrawal2012further, agrawal2012analysis}
provide regret bounds for posterior sampling when applied to
MAB problems with finitely many independent actions and rewards that follow Bernoulli
processes.  These results demonstrate that posterior sampling is asymptotically optimal 
for the class of problems considered. 
A key feature of the bounds is their dependence on the difference between the optimal
and second-best mean-reward values.  Such bounds tend not to be meaningful when the number of actions is large or infinite
unless they can be converted to bounds that are independent of this gap, which is sometimes the case.

In this paper, we establish distribution-independent bounds. When the action space
 $\mathcal{A}$ is finite, we establish a finite time Bayesian regret bound of order 
$\sqrt{\left|\mathcal{A}\right|T\log T}$.  This matches what is implied by the analysis 
of \citet{agrawal2012further}.  However, our bound does not require actions are modeled independently, and our approach also leads to meaningful bounds for problems 
with large or infinite action sets. 

Only one other paper has studied posterior sampling in a context
involving dependent actions \cite{agrawal2013linear}.  That paper considers a
contextual bandit model with arms whose mean-reward values
are given by a $d$-dimensional linear model.  The cumulative $T$-period regret is shown
to be of order $\frac{d^2}{\epsilon}\sqrt{T^{1+\epsilon}}\ln(Td)\ln\frac{1}{\delta}$
with probability at least $1-\delta$. Here $\epsilon\in(0,1)$ is
a parameter used by the algorithm to control how quickly the posterior
distribution concentrates.  The Bayesian regret bounds we will establish are stronger than those 
implied by the results of \citet{agrawal2013linear}. In particular, we provide a Bayesian regret 
bound of order $d\sqrt{T}\ln T$ that holds for any compact set of actions. This is order--optimal
up to a factor of $\ln T$ \cite{rusmevichientong2010linearly}.

We are also the first to establish finite time performance bounds for several other
problem classes.  One applies to linear models when the vector of coefficients 
is likely to be sparse; this bound is stronger than the aforementioned one that applies
to linear models in the absence of sparsity assumptions.  We establish the the first bounds
for posterior sampling when applied to generalized linear models and to problems with a 
general Gaussian prior. Finally, we establish bounds that apply very broadly and depend 
on a new notion of dimension.

Unlike most of the relevant literature, we study MAB problems in a general framework, allowing for complicated relationships between the
rewards generated by different actions. The closest related work is that of \citet{aminbandits}, who consider the problem of 
learning the optimum of a function that lies in a known, but otherwise arbitrary set of functions. 
They provide bounds based on a new notion of dimension, but unfortunately this notion does not provide a
bound for posterior sampling. 
Other work provides general bounds for contextual bandit problems where the context space is allowed
to be infinite, but the action space is small (see, e.g., \cite{dudik2011efficient}). Our model captures contextual bandits as 
a special case, but we emphasize problem instances with large or infinite action sets, and where the goal is to learn 
without sampling every possible action.

A focus of our paper is the connection between posterior sampling and UCB approaches.
We discuss UCB algorithms in some detail in Section \ref{sec: algorithms}.  UCB algorithms have 
been the primary approach considered in the segment of the stochastic MAB literature that treats
models with dependent arms.  Other approaches are the knowledge gradient algorithm 
 \cite{ryzhov2012knowledge}, forced exploration schemes for linear bandits 
\cite{abbasi2009forced, rusmevichientong2010linearly, deshpande2012linear}, and exponential-weighting 
schemes \cite{dudik2011efficient}. 

There is an immense and rapidly growing literature on bandits with independent arms and on adversarial bandits. 
Theoretical work on stochastic bandits with independent arms often focuses on 
UCB algorithms \cite{lai1985asymptotically, auer2002finite} or on the Gittin's index approach \cite{gittins1979dynamic}. 
\citet{bubeck2012regret} provide a review of work on UCB algorithms and on adversarial bandits. \citet{gittins2011multi} 
cover work on Gittin's indices and related extensions.

Since an initial version of this paper was made publicly available, the literature on the analysis of posterior sampling has rapidly grown. 
\citet{Korda2013Thompson} extend their earlier work \cite{kaufmann2012thompson} to the case where reward distributions lie in the 
1--dimensional exponential family. \citet{bubeck2013prior} combine the regret decomposition we derive in Section \ref{sec:implicit} with the confidence bound analysis of \citet{audibert2009minimax}
to tighten the bound provided in Section \ref{subsec: finite}, and also consider a problem setting where the regret of posterior sampling is bounded uniformly over time. \citet{li2013generalized} 
explores a connection between posterior sampling and exponential weighting schemes, and \citet{gopalan2013thompson} study the asymptotic growth rate of regret in problems with dependent arms.

\section{Problem Formulation.}
\label{sec:formulation}

We consider a model involving a set of actions $\mathcal{A}$ and a set of real-valued functions 
$\mathcal{F}=\left\{ f_\rho:\mathcal{A} \mapsto \mathbb{R} \right | \rho \in \Theta\}$, indexed by a parameter that
takes values from an index set $\Theta$.  We will define random variables with respect to a probability 
space $(\Omega, \mathbb{F}, \mathbb{P})$.  A random variable $\theta$ indexes the true reward
function $f_\theta$.  At each time $t$, the agent 
is presented with a possibly random subset $\mathcal{A}_t \subseteq \mathcal{A}$ and selects an 
action $A_t \in \mathcal{A}_t$, after which she observes a reward $R_t$.  

We denote by $H_t$ the history $(\mathcal{A}_1, A_1, R_{1}, \ldots, \mathcal{A}_{t-1}, A_{t-1}, R_{t-1}, \mathcal{A}_t)$
of observations available to the agent when choosing an action $A_t$.
The agent employs a policy $\pi = \{\pi_t | t \in \mathbb{N}\}$, which is a deterministic sequence
of functions, each mapping the history $H_t$ to a probability distribution over actions
$\mathcal{A}$.  For each realization of $H_t$, $\pi_t(H_t)$ is a distribution over $\mathcal{A}$ with support $\mathcal{A}_t$,
though with some abuse of notation, we will often write this distribution as $\pi_t$. 
The action $A_t$ is selected by sampling from the distribution $\pi_t$, so that 
$\mathbb{P}(A_t \in \cdot | \pi_t) = \mathbb{P}(A_t \in \cdot | H_t) = \pi_t(\cdot)$.
We assume that $\mathbb{E}[R_t | H_t, \theta, A_t] = f_\theta(A_t)$. In other words,
the realized reward is the mean-reward value corrupted by zero-mean noise.  We will also assume
that for each $f \in \mathcal{F}$ and $t \in \mathbb{N}$, $\arg\max_{a \in \mathcal{A}_t} f(a)$ is nonempty with probability one,
though algorithms and results can be generalized to handle cases where this assumption does not hold.

The $T$-period regret of a policy $\pi$ is the random variable defined by 
$$\text{Regret}\left(T,\,\pi, \theta \right)=\sum_{t=1}^{T}\mathbb{E}\left[\left.\max_{a\in\mathcal{A}_{t}}f_{\theta}(a)-f_{\theta}\left(A_t\right) 
\ \right\vert \   \theta  \right].$$
The $T$-period Bayesian regret is defined by $\mathbb{E}\left[\text{Regret}\left(T,\,\pi, \theta \right)\right]$, where the expectation is taken with respect to 
the prior distribution over $\theta$.    Hence, 
\[
\text{BayesRegret}\left(T,\,\pi\right) = \sum_{t=1}^{T} \mathbb{E}\left[\max_{a\in\mathcal{A}_{t}}f_{\theta}(a)-f_{\theta}\left(A_{t}\right)\right].
\]
This quantity is also called Bayes risk, or simply expected regret. 
\begin{remark}
Measurability assumptions are required for the above expectations to be well-defined.
In order to avoid technicalities that do not present fundamental obstacles in the contexts we consider, we will not explicitly address
measurability issues in this paper and instead simply assume that functions under consideration satisfy conditions that ensure relevant 
expectations are well-defined.
\end{remark}
\begin{remark}
All equalities between random variables in this paper  hold almost surely with respect to the underlying probability space. 
\end{remark}

\subsection{On Regret and Bayesian Regret.}

To interpret results about the regret and Bayesian regret of various algorithms and to appreciate their practical implications, it is useful to 
take note of several properties of and relationships between these performance measures.  For starters, 
asymptotic bounds on Bayesian regret are essentially asymptotic bounds on regret. In particular, if $\mathrm{BayesRegret}(T,\pi)=O(g(T))$ for some non-negative 
function $g$ then an application of Markov's inequality shows $\mathrm{Regret}(T,\pi, \theta)=O_P(g(T))$. 
Here $O_P$ indicates that $\mathrm{Regret}(T,\pi, \theta)/g(T)$ is stochastically bounded under the prior distribution. In other words, for all $\epsilon>0$ there exists $M>0$ such that $$\mathbb{P}\left( \frac{\mathrm{Regret}(T,\pi, \theta)}{g(T)} \geq M \right) \leq \epsilon \qquad \forall T \in \mathbb{N}.$$
This observation can be further extended to establish a sense in which Bayesian regret is robust to prior mis-specification.  
In particular, if the agent's prior over $\theta$ is $\mu$ but for convenience he selects actions as though his prior were an alternative $\tilde{\mu}$, the resulting Bayesian regret satisfies
$$\mathbb{E}_{\theta_0 \sim \mu}\left[\mathrm{Regret}(T,\pi,\theta_0) \right] \leq \left\| \frac{d\mu}{d\tilde{\mu}} \right\|_{\tilde{\mu},\infty}   \mathbb{E}_{\theta_0 \sim \tilde{\mu}}\left[\mathrm{Regret}(T,\pi,\theta_0)\right],$$
where $d\mu/d\tilde{\mu}$ is the Radon-Nikodym derivative\footnote{Note that the Radon-Nikodym derivative is only well defined when $\mu$ is absolutely continuous with respect to $\tilde{\mu}$.}   of $\mu$ with respect to $\tilde{\mu}$ and $\|\cdot\|_{\tilde{\mu},\infty}$ is the essential supremum magnitude with respect to $\tilde{\mu}$.  
Note that the final term on the right-hand-side is the Bayesian regret for a problem with prior $\tilde{\mu}$ without mis-specification.

It is also worth noting that an algorithm's Bayesian regret can only differ significantly from its worst-case regret if regret varies significantly depending on the realization of $\theta$. This provides one method of converting Bayesian regret bounds to regret bounds. For example, consider the linear model $f_\theta(a)=\theta^T \phi(a)$ where $\Theta = \left\{ \rho\in \mathbb{R}^d : \left\Vert \rho \right\Vert_2=S \right\}$ is the boundary of a hypersphere in $\mathbb{R}^d$.  Let $\mathcal{A}_t=\mathcal{A}$ for each $t$ and let the set of feature vectors be $\left\{ \phi(a) \vert a \in \mathcal{A}\right\}=\left\{u\in \mathbb{R}^d \vert \left\Vert u \right\Vert_2 \leq 1 \right\}.$  Consider a problem instance where $\theta$ is uniformly distributed  over $\Theta$, and the noise terms $R_t-f_\theta(A_t)$ are independent of $\theta$. By symmetry, the regret of most reasonable algorithms for this problem should be the same for all realizations of $\theta$, and indeed this is the case for posterior sampling. Therefore, in this setting Bayesian regret is equal to worst--case regret. This view also suggests that in order to attain strong minimax regret bounds, one should not choose a uniform prior as in \citet{agrawal2012further}, but should instead place more prior weight on the worst possible realizations of $\theta$ (see the discussion of ``least favorable'' prior distributions in \citet{lehmann1998theory}).

\subsection{On Changing Action Sets.}

Our stochastic model of action sets $\mathcal{A}_{t}$ is distinct relative to most of
the multi-armed bandit literature, which assumes that $\mathcal{A}_t = \mathcal{A}$.  
This construct allows our formulation to address
a variety of practical issues that are usually viewed as beyond the scope of standard multi-armed 
bandit formulations.  Let us provide three examples.

\begin{example}
{\bf Contextual Models.} The contextual multi-armed bandit model is a special case of the 
formulation presented above.  In such a model, an exogenous Markov process $X_{t}$
taking values in a set $\mathcal{X}$ influences rewards.  In particular, the expected reward
at time $t$ is given by $f_{\theta}(a,X_{t})$.  However, this is mathematically equivalent to a problem 
with stochastic time-varying decision sets $\mathcal{A}_{t}$. In particular, one can define the
set of actions to be the set of state-action pairs $\mathcal{A}:=\left\{ \left(x,\, a\right):\, x\in\mathcal{A},\, a\in\mathcal{A}(x)\right\} $,
and the set of available actions to be $\mathcal{A}_{t}=\left\{ \left(X_{t},\, a\right):\, a\in\mathcal{A}(X_{t})\right\} $.
\end{example}

\begin{example}
{\bf Cautious Actions.}
In some applications, one may want to explore without risking terrible
performance. This can be accomplished by restricting the set
$\mathcal{A}_{t}$ to conservative actions.  
Then, the instantaneous regret in our framework
is the gap between the reward from the chosen action and the reward
from the best conservative action.  In many settings, the 
Bayesian regret bounds we will establish for posterior sampling imply that 
the algorithm either attains near-optimal
performance or converges to a point where any better decision is
unacceptably risky.

A number of formulations of this flavor are amenable to efficient implementations of Posterior Sampling.
For example, consider a problem where $\mathcal{A}$ is a polytope or ellipsoid in $\mathbb{R}^d$ and
$f_{\theta}(a)=\left\langle a,\,\theta\right\rangle $.
Suppose $\theta$ has a Gaussian prior and that reward noise is Gaussian.  Then, the posterior distribution of $\theta$ is Gaussian.
Consider an ellipsoidal confidence set 
$\mathcal{U}_{t}=\left\{ u\mid\left\Vert u-\mu_{t}\right\Vert _{\Sigma_{t}}\leq\beta\right\}$,
for some scalar constant $\beta > 0$,
where $\mu_t$ and $\Sigma_t$ are the mean and covariance matrix of $\theta$,
conditioned on $H_{t}$.  
One can attain good worst-case performance with high probability
by solving the robust optimization problem $V_{\rm robust} \equiv \max_{a\in\mathcal{A}}\min_{u\in\mathcal{U}_t}\left\langle a,u\right\rangle$, 
which is a tractable linear saddle-point problem.  Letting our cautious set be given by 
\[
\mathcal{A}_{t}=\left\{ a\in\mathcal{A}\mid\min_{u\in\mathcal{U}_{t}}\left\langle a,u\right\rangle \geq V_{\rm robust}-\alpha\right\} 
\]
for some scalar constant $\alpha > 0$, we can then select an optimal cautious action given $\theta$ by solving $\max_{a \in \mathcal{A}_{t}} \left\langle a,\theta \right\rangle$, which is equivalent to
$$
\begin{array}{ll}
\text{\rm maximize} & \left\langle a,\theta\right\rangle \\
\text{\rm subject to} & a\in\mathcal{A}\\
 & \left\Vert a\right\Vert _{\Sigma_{t}^{-1}}\leq\frac{1}{\beta}\left(\left\langle a,\mu_{t}\right\rangle -V_{\rm robust}+\alpha\right).
 \end{array}
$$
This problem is computationally tractable,
which accommodates efficient implementation of posterior sampling.
\end{example}

\begin{example}
{\bf Adaptive Adversaries.}
Consider a model in which rewards are influenced by the choices of an adaptive adversary. 
At each time period, the adversary selects an action $A^-_t$ from some set 
$\mathcal{A}^-$ based on past observations. The agent observes this action, responds
with an action $A^+_t$ selected from a set $\mathcal{A}^+$, 
and receives a reward that depends on the pair of actions $(A_{t}^{+}, A_{t}^{-})$.  This fits our framework if the action $A_t$ is taken to be the pair $(A^+_t,A^-_t)$,
and the set of actions available to the agent is $\mathcal{A}_t = \{(a,A^-_t) | a \in \mathcal{A}^+\}$.
\end{example}

\section{Algorithms.}
\label{sec: algorithms}

We will establish finite time performance bounds for posterior sampling by leveraging prior results pertaining to UCB algorithms 
and a connection we will develop between the two classes of algorithms.  To set the stage for our 
analysis, we discuss the algorithms in this section.
 
\subsection{UCB Algorithms.}
\label{subsec: ucb}

UCB algorithms have received a great deal of attention in the MAB literature.  Such an algorithm
makes use of a sequence of upper confidence bounds $U = \{U_t | t \in \mathbb{N}\}$, each of which 
is a function that takes the history $H_t$ as its argument.
For each realization of $H_t$, $U_t(H_t)$ is a function mapping $\mathcal{A}$ to $\mathbb{R}$.
With some abuse of notation, we will often write this function as $U_t$ and its value at $a \in \mathcal{A}$
as $U_t(a)$.  The upper confidence bound $U_{t}(a)$ represents the greatest value of $f_{\theta}(a)$ that 
is statistically plausible given $H_t$.  A UCB algorithm
selects an action $\bar{A_{t}} \in \arg\max_{a\in\mathcal{A}_t}U_{t}(a)$
that maximizes the upper confidence bound.  We will assume that the argmax operation breaks ties among 
optima in a deterministic way.  As such, each action is determined by the history $H_t$,
and for the policy $\pi = \{\pi_t | t \in \mathbb{N}\}$ followed by a UCB algorithm, each action 
distribution $\pi_t$ concentrates all probability on a single action.

As a concrete example, consider Algorithm \ref{alg:independentUCB}, proposed by \citet{auer2002finite} to 
address MAB problems with a finite number of independent actions.  For such problems, $\mathcal{A}_t = \mathcal{A}$, 
$\theta$ is a vector with one independent component per action, and the reward
function is given by $f_\theta(a) = \theta_a$.
The algorithm begins by selecting each action once.
Then, for each subsequent time $t > |\mathcal{A}|$, the algorithm generates point estimates of action rewards,
defines upper confidence bounds based on them, and selects actions accordingly.  For each action $a$, the 
point estimate $\hat{\theta}_t(a)$ is taken to be the average reward obtained from samples of action $a$ taken
prior to time $t$.  The upper confidence bound is produced by adding an ``uncertainty bonus'' $\beta \sqrt{\log t / N_t(a)}$
to the point estimate, where $N_t(a)$ is the number of times action $a$ was selected prior to time $t$ and 
$\beta$ is an algorithm parameter generally selected based on reward variances.
This uncertainty bonus leads to an optimistic assessment of expected reward when
there is uncertainty, and it is this optimism that encourages exploration that reduces uncertainty.  As 
$N_t(a)$ increases, uncertainty about action $a$ diminishes
and so does the uncertainty bonus.  The $\log t$ term ensures that the agent does not permanently
rule out any action, which is important as there is always some chance of obtaining an overly pessimistic
estimate by observing an unlikely sequence of rewards.

Our second example treats a linear bandit problem.  Here we assume $\theta$ is drawn from a 
normal distribution $N(\mu_0, \Sigma_0)$ but without assuming that the covariance matrix
is diagonal.  We consider a linear reward function $f_\theta(a) = \langle \phi(a), \theta\rangle$ and 
assume the reward noise 
$R_t-f_\theta(A_t)$ is normally distributed and independent
from $(H_t, A_t, \theta)$.  One can show that, conditioned on the history $H_t$, $\theta$
remains normally distributed.  Algorithm \ref{alg:Linear--Gaussian UCB}
presents an implementation of UCB algorithm for this problem.  The expectations can be computed 
efficiently via Kalman filtering. The algorithm employs upper confidence bound $ \left\langle \phi(a),\,\mu_{t}\right\rangle+ \beta \log(t) \left\| \phi(a) \right\|_{\Sigma_t }$.
The term $ \left\| \phi(a) \right\|_{\Sigma_t}$  captures the posterior variance of $\theta$ in the direction $\phi(a)$, and, 
as with the case of independent arms, causes the uncertainty bonus  $\beta \log(t) \left\| \phi(a) \right\|_{\Sigma_t }$ to diminish as the number of observations increases. 

\begin{figure}
\begin{minipage}[t]{3.15in}

\algsetup{indent=2em}
\begin{algorithm}[H]
\caption{Independent UCB}\label{alg:independentUCB}
\begin{algorithmic}[1]
\STATE \textbf{Initialize}: Select each action once \protect\\
\STATE \textbf{Update Statistics}: For each $a \in \mathcal{A}$, \\
$\hat{\theta}_t(a) \leftarrow$  sample average of observed rewards \protect\\
$N_t(a) \leftarrow$  number of times $a$ sampled so far \protect\\
\STATE \textbf{Select Action}: \protect\\
$\overline{A}_{t}\underset{a\in\mathcal{A} }{\arg\max}\left\{ \hat{\theta}_{t}(a)+ \beta \sqrt{\frac{\log t}{N_{t}(a)}} \right\} $ \protect\\
\STATE \textbf{Increment $t$ and Goto Step 2}\\
\end{algorithmic}
\end{algorithm}

\end{minipage}
\hspace{0.5cm}
\begin{minipage}[t]{3.15in}
\centering

\algsetup{indent=2em}
\begin{algorithm}[H]
\caption{Linear--Gaussian UCB}\label{alg:Linear--Gaussian UCB}
\begin{algorithmic}[1]
\STATE \textbf{Update Statistics}:  \\
$\mu_t \leftarrow \mathbb{E}[\theta | H_t]$ \protect\\
$\Sigma_t \leftarrow \mathbb{E}[(\theta-\mu_t)(\theta-\mu_t)^\top | H_t]$ \protect\\
\STATE \textbf{Select Action}: \\
$\overline{A}_{t}\in \underset{a\in\mathcal{A} }{\arg\max}\left\{ \left\langle \phi(a),\,\mu_{t}\right\rangle+ \beta \log(t) \left\| \phi(a) \right\|_{\Sigma_t }\right\} $ \protect\\
\STATE \textbf{Increment $t$ and Goto Step 1}\\
\end{algorithmic}
\end{algorithm}
\end{minipage}

\end{figure}

\subsection{Posterior Sampling.}
\label{subsec: posterior}

The posterior sampling algorithm simply samples each action according to the probability it is optimal.
In particular, the algorithm applies action sampling distributions $\pi_t = \mathbb{P}\left(A_t^* \in \cdot \mid H_{t}\right)$,
where $A^*_t$ is a random variable that satisfies $A^*_t \in \arg\max_{a \in \mathcal{A}_t} f_\theta(a)$.
Practical implementations typically operate by, at each time $t$, sampling an index $\hat{\theta}_t \in \Theta$
from the distribution $\mathbb{P}\left(\theta \in \cdot \mid H_{t}\right)$ and then generating
an action $A_t \in \arg\max_{a \in \mathcal{A}_t} f_{\hat{\theta}_t}(a)$.
To illustrate, let us provide concrete examples that address problems analogous to
Algorithms \ref{alg:independentUCB} and \ref{alg:Linear--Gaussian UCB}.

Our first example involves a model with independent arms.  In particular, 
suppose $\theta$ is drawn from a normal distribution $N(\mu_0, \Sigma_0)$ with a diagonal covariance
matrix $\Sigma_0$, the reward function
is given by $f_\theta(a) = \theta_a$, and the reward noise 
$R_t-f_\theta(A_t)$ is normally distributed and independent
from $(H_t, A_t, \theta)$.  It then follows that, conditioned on the history $H_t$, $\theta$
remains normally distributed with independent components.  Algorithm \ref{alg:independentPosteriorSampling}
presents an implementation of posterior sampling for this problem.  The expectations are easy
to compute and can also be computed recursively.

Our second example treats a linear bandit problem.  Algorithm \ref{alg:linearPosteriorSampling}
presents a posterior sampling analogue to Algorithm  \ref{alg:Linear--Gaussian UCB}.  As before, we assume $\theta$ is drawn from a 
normal distribution $N(\mu_0, \Sigma_0)$. We consider a linear reward function $f_\theta(a) = \langle \phi(a), \theta\rangle$ and 
assume the reward noise $R_t-f_\theta(A_t)$ is normally distributed and independent from $(H_t, A_t, \theta)$.  
%

\begin{figure}[H]
\begin{minipage}[t]{3.15in}
\algsetup{indent=2em}
\begin{algorithm}[H]
\caption{\protect\\ Independent Posterior Sampling}
\label{alg:independentPosteriorSampling}
\begin{algorithmic}[1]
\STATE \textbf{Sample Model}: \\
$\hat{\theta}_t \sim N(\mu_{t-1}, \Sigma_{t-1})$
\STATE \textbf{Select Action}: \protect\\
$A_{t}\in\arg\max_{a\in\mathcal{A}} \hat{\theta}_{t}(a)$ \protect\\
\STATE \textbf{Update Statistics}: For each $a$,  \\
$\mu_{ta} \leftarrow \mathbb{E}[\theta_a | H_t]$ \protect\\
$\Sigma_{taa} \leftarrow \mathbb{E}[(\theta_a-\mu_{ta})^2| H_t]$ \protect\\
\STATE \textbf{Increment $t$ and Goto Step 1}\\
\end{algorithmic}
\end{algorithm}

\end{minipage}
\hspace{0.5cm}
\begin{minipage}[t]{3.15in}
\centering

\algsetup{indent=2em}
\begin{algorithm}[H]
\caption{\protect\\ Linear Posterior Sampling}
\label{alg:linearPosteriorSampling}
\begin{algorithmic}[1]
\STATE \textbf{Sample Model}: \\
$\hat{\theta}_t \sim N(\mu_{t-1}, \Sigma_{t-1})$
\STATE \textbf{Select Action}: \protect\\
$A_{t}\in\arg\max_{a\in\mathcal{A}} \langle\phi(a), \hat{\theta}_{t}\rangle$ \protect\\
\STATE \textbf{Update Statistics}:  \\
$\mu_t \leftarrow \mathbb{E}[\theta | H_t]$ \protect\\
$\Sigma_t \leftarrow \mathbb{E}[(\theta-\mu_t)(\theta-\mu_t)^\top | H_t]$ \protect\\
\STATE \textbf{Increment $t$ and Goto Step 1}\\
\end{algorithmic}
\end{algorithm}
\end{minipage}
\end{figure}

\subsection{Potential Advantages of Posterior Sampling.}\label{subsec:advantages}

Well designed optimistic algorithms can be extremely effective. When simple and efficient UCB algorithms are available, 
they may be preferable to posterior sampling. Our work is primarily motivated by the challenges in designing optimistic algorithms
to address very complicated problems, and the important advantages posterior sampling sometimes offers in such cases. 

The performance of a UCB algorithm depends critically on the choice of upper confidence bounds  $\left( U_{t} : t\in \mathbb{N}\right)$. These functions should be chosen so that $U_t(A^*) \geq f_{\theta}(A^*)$ with high probability.
However, unless the posterior distribution of $f_{\theta}(a)$ can be expressed in closed form, computing high quantiles of this distribution can 
require extensive Monte Carlo simulation for each possible action. 
In addition, since $A^*$ depends on $\theta$,  
it isn't even clear that $U_t(a)$ should be set to a particular quantile of the posterior distribution of $f_{\theta}(a)$. 
Strong frequentist upper confidence bounds were recently developed \cite{KL-UCB2013} for problems with independent arms, 
but it is often unclear how to generate tight confidence sets for more complicated models. 
In fact, even in the case of a linear model, the design of confidence sets has required sophisticated tools
from the study of multivariate self-normalized martingale processes \cite{abbasi2011improved}. We believe 
posterior sampling provides a powerful tool for practitioners, as a posterior sampling approach can be designed 
for complicated models without sophisticated statistical analysis. Further, posterior sampling does not require
computing posterior distributions but only sampling from posterior distributions.  As such, Markov
chain Monte Carlo methods can often be used to efficiently generate samples even when the posterior
distribution is complex.

Posterior sampling can also offer critical computational advantages over UCB algorithms when the action space 
is intractably large.  Consider the problem of learning to solve a linear program, where the action set $\mathcal{A}$ is a polytope encoded in terms of linear inequalities,  and $f_{\theta}(a):=\langle\phi(a), \theta \rangle$ is a linear function.  Algorithm \ref{alg:Linear--Gaussian UCB} becomes impractical, because, as observed by \citet{dani2008stochastic},
the action selection step entails solving a problem equivalent to 
linearly constrained negative definite quadratic optimization, which is NP hard \cite{sahni1974}.\footnote{\citet{dani2008stochastic} studies a slightly different UCB algorithm, but the optimization step shares the same structure.} 
By contrast, the action selection step of Algorithm \ref{alg:linearPosteriorSampling} only requires solving a linear program. 
The figure below displays the level sets of the linear objective $\langle\phi(a), \hat{\theta} \rangle$ and of the upper confidence bounds used by Algorithm  \ref{alg:Linear--Gaussian UCB}. While posterior sampling preserves the  linear structure of the functions $f_{\theta}(a)$, it is challenging to maximize the upper confidence bounds of Algorithm  \ref{alg:Linear--Gaussian UCB}, which are strictly convex. 

\begin{figure}[h!]
\centering
\begin{subfigure}{.5\textwidth}
  \centering
  \includegraphics[width=1\linewidth]{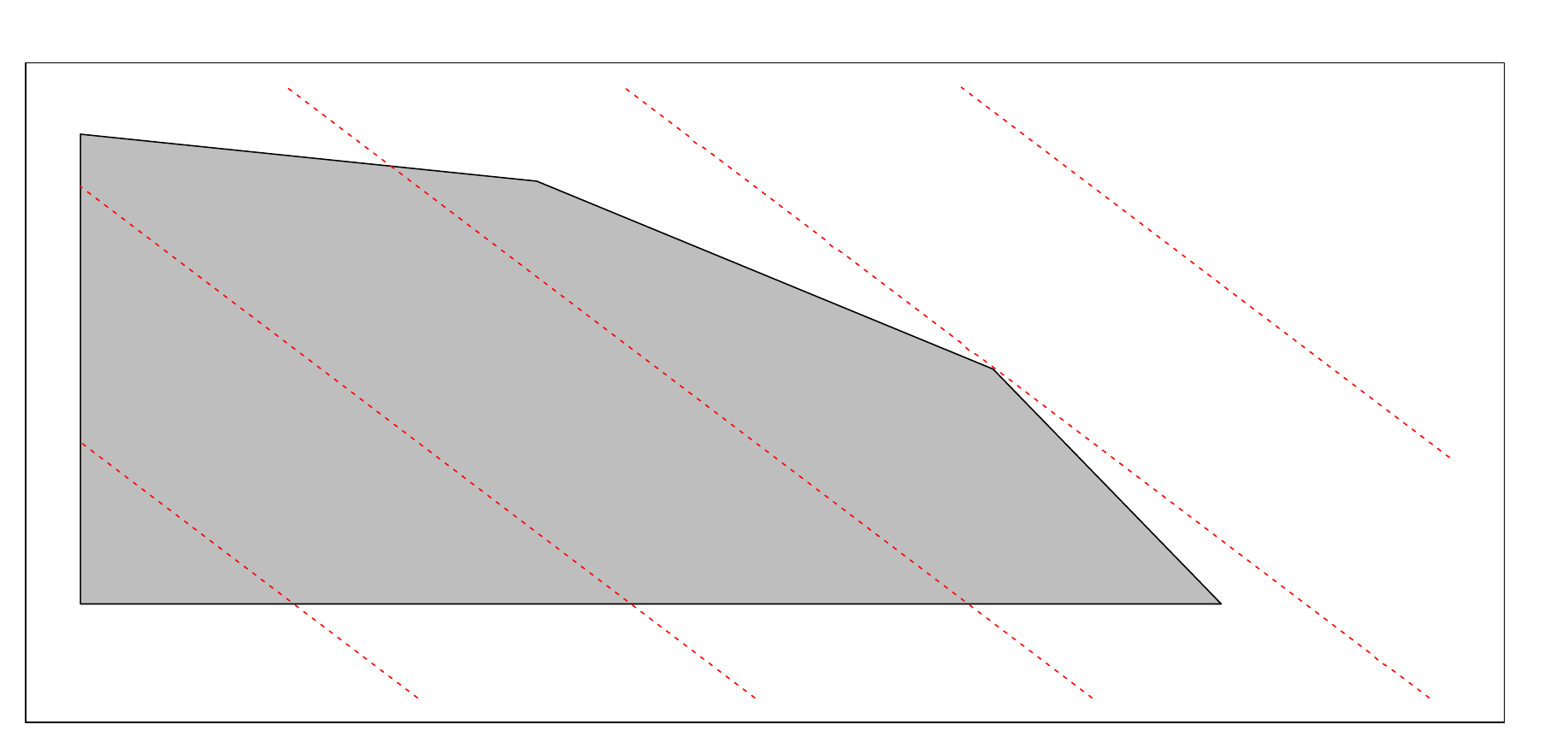}
  \caption{{\it Tractable} problem of maximizing a linear \\function over a polytope}
  \label{fig:sub1}
\end{subfigure}%
\begin{subfigure}{.5\textwidth}
  \centering
  \includegraphics[width=1\linewidth]{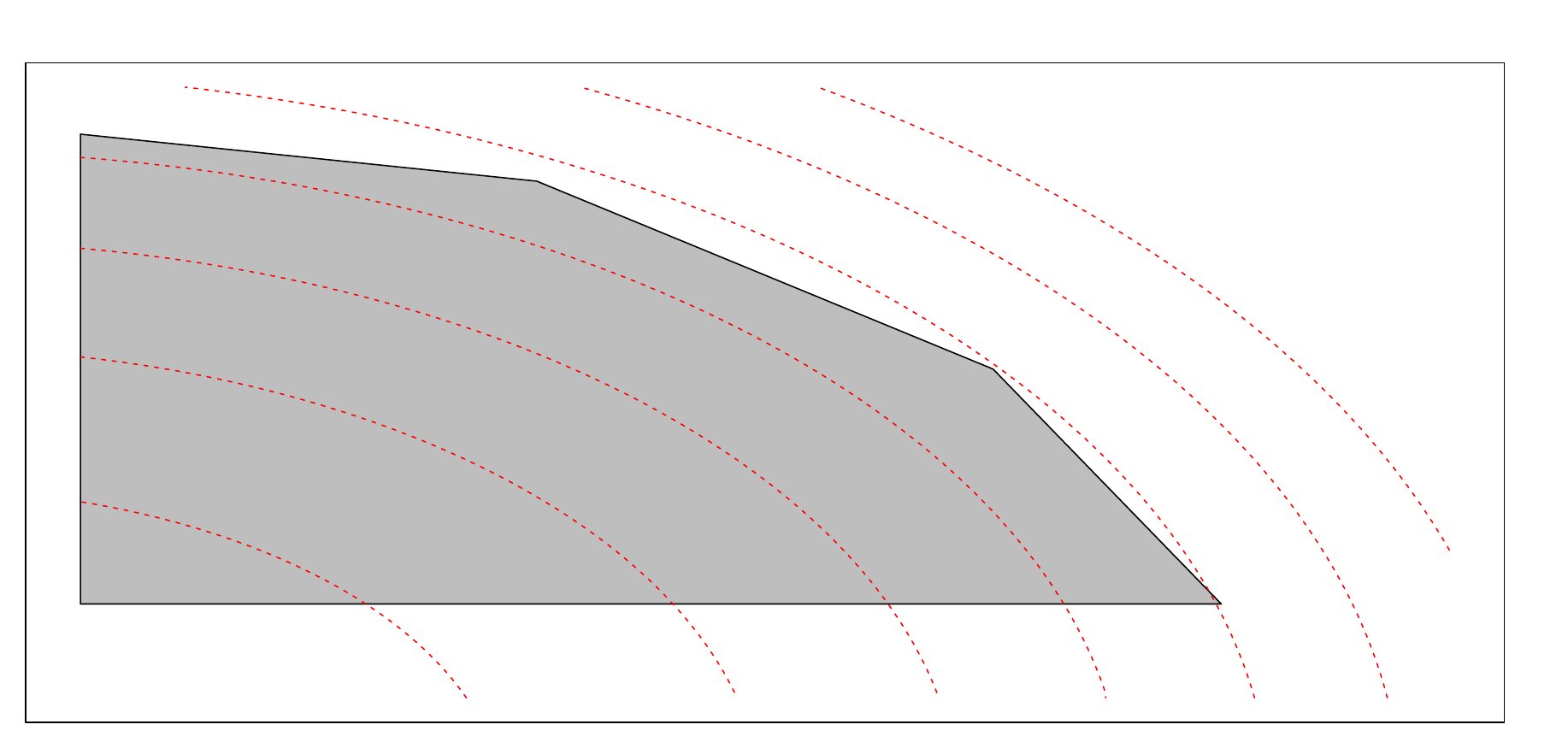}
  \caption{{\it Intractable} problem of maximizing an upper \\confidence bound over a polytope}
  \label{fig:sub2}
\end{subfigure}
\label{fig:polytope}
\end{figure}

It is worth mentioning that because posterior sampling requires specifying a fully probabilistic model of the underlying system, it may not be ideally suited for every practical setting.  In particular, when dealing with some complex classes of functions, specifying an appropriate prior can be a challenge while there may be alternative algorithms that address the problem in an elegant and practically effective way.

\section{Confidence Bounds and Regret Decompositions.}
\label{sec:implicit}

Unlike UCB algorithms, posterior sampling does not make use of upper confidence bounds to encourage
exploration and instead relies on randomization.  As such, the two classes of algorithm seem very different.
However, we will establish in this section a connection that will enable us in Section \ref{sec:UCB} to 
derive performance bounds for posterior sampling from those that apply to UCB algorithms.  Since
UCB algorithms have received much more attention, this leads to a number of new
results about posterior sampling.  Further, the relationship yields insight into 
the performance advantages of posterior sampling.

\subsection{UCB Regret Decomposition.}

Consider a UCB algorithm with an upper confidence bound sequence $U = \{U_t | t \in \mathbb{N}\}$.
Recall that $\bar{A}_t \in \arg\max_{a \in \mathcal{A}_t} U_t(a)$ and $A_t^* \in \arg\max_{a \in \mathcal{A}_t} f_\theta(a)$.
We have the following simple regret decomposition: 
\begin{eqnarray}
f_{\theta}\left(A_{t}^{*}\right)-f_{\theta}\left(\bar{A_{t}}\right) & = & f_{\theta}\left(A_{t}^{*}\right)-U_{t}(\bar{A_{t}})+U_{t}(\bar{A_{t}})-f_{\theta}\left(\bar{A_{t}}\right)\nonumber \\
 & \leq & \left[f_{\theta}\left(A_{t}^{*}\right)-U_{t}(A_{t}^{*})\right]+\left[U_{t}(\bar{A_{t}})-f_{\theta}\left(\bar{A_{t}}\right)\right].\label{eq: UCB Regret Decomposition}
\end{eqnarray}
The inequality follows from the fact that $\bar{A_{t}}$ is chosen to maximize
$U_{t}$. If the upper confidence bound is an upper bound with high probability,
as one would expect from a UCB algorithm,
then the first term is negative with high probability. 
The second term, $U_{t}(\bar{A_{t}})-f_{\theta}\left(\bar{A_{t}}\right)$, 
penalizes for the width of the confidence interval.
As actions are sampled $U_{t}$ should diminish
and converge on $f_{\theta}$.  As such, both terms of the decomposition should 
eventually vanish.  An important feature of this decomposition is that,
so long as the first term is negative, it bounds
regret in terms of uncertainty about \textit{the current action} $\bar{A}_t$. 

Taking the expectation of (\ref{eq: UCB Regret Decomposition}) establishes
that the $T$-period Bayesian regret of a UCB algorithm satisfies
\begin{equation}
\text{BayesRegret}\left(T,\,\pi^U \right) \leq \mathbb{E}\sum_{t=1}^{T}\left[U_{t}(\bar{A_{t}})-f_{\theta}(\bar{A_{t}})\right]+\mathbb{E}\sum_{t=1}^{T}\left[f_{\theta}\left(A_{t}^{*}\right)-U_{t}(A_{t}^{*})\right], \label{eq: UCB Risk Decomposition}
\end{equation}
where $\pi^U$ is the policy derived from $U$.

\subsection{Posterior Sampling Regret Decomposition.}
As established by the following proposition, the Bayesian regret of posterior sampling decomposes in a way analogous to what 
we have shown for UCB algorithms. Recall that, with some abuse of notation, for an upper confidence bound sequence $\{U_t | t \in \mathbb{N}\}$
we denote by $U_t (a)$ the random variable $U_t(H_t)(a)$. The following proposition allows $U_{t}$ to be an arbitrary real valued function of $H_{t}$ and $a\in \mathcal{A}$. Let $\pi^{\rm PS}$ denote the policy followed by posterior sampling.
\begin{proposition}
For any upper confidence bound sequence $\{U_t | t \in \mathbb{N}\}$,
\begin{equation}
\textup{BayesRegret}\left(T,\,\pi^{\rm PS}\right) = \mathbb{E}\sum_{t=1}^{T}\left[U_{t}(A_{t})-f_{\theta}(A_{t})\right]+\mathbb{E}\sum_{t=1}^{T}\left[f_{\theta}\left(A_{t}^{*}\right)-U_{t}(A_{t}^{*})\right],\label{eq: posterior sampling Risk Decomposition}
\end{equation}
for all $T \in \mathbb{N}$.
\end{proposition}
\proof{\textup{Proof.}}
Note that, conditioned on $H_t$, the optimal action $A^*_t$ and the action $A_t$ selected by posterior sampling are identically distributed,
and $U_t$ is deterministic.
Hence, $\mathbb{E}\left[U_{t}(A_{t}^{*})\mid H_t\right]=\mathbb{E}\left[U_{t}(A_{t})\mid H_t\right]$.  Therefore
\begin{eqnarray*}
\mathbb{E}\left[f_{\theta}(A_{t}^{*})-f_{\theta}\left(A_{t}\right)\right] & = & \mathbb{E}\left[\mathbb{E}\left[f_{\theta}(A_{t}^{*})-f_{\theta}\left(A_{t}\right)\mid H_{t}\right]\right]\\
 & = & \mathbb{E}\left[\mathbb{E}\left[U_{t}\left(A_{t}\right)-U_{t}\left(A_{t}^{*}\right)+f_{\theta}(A_{t}^{*})-f_{\theta}\left(A_{t}\right)\mid H_{t}\right]\right]\\
 & = & \mathbb{E}\left[\mathbb{E}\left[U_{t}(A_{t})-f_{\theta}(A_{t})\mid H_{t}\right]+\mathbb{E}\left[f_{\theta}\left(A_{t}^{*}\right)-U_{t}(A_{t}^{*})\mid H_{t}\right]\right]\\
 & = & \mathbb{E}\left[U_{t}(A_{t})-f_{\theta}(A_{t})\right]+\mathbb{E}\left[f_{\theta}\left(A_{t}^{*}\right)-U_{t}(A_{t}^{*})\right].
\end{eqnarray*}
Summing over $t$ gives the result. \Halmos
\endproof

To compare (\ref{eq: UCB Risk Decomposition}) and (\ref{eq: posterior sampling Risk Decomposition})
consider the case where $f_{\theta}$ takes values in $[0,\, C]$.  Then,
$$\text{BayesRegret}\left(T,\,\pi^U\right) \leq \mathbb{E}\sum_{t=1}^{T}\left[U_{t}(\bar{A_{t}})-f_{\theta}(\bar{A_{t}})\right]
+ C\sum_{t=1}^{T}\mathbb{P}\left(f_{\theta}\left(A_{t}^{*}\right)>U_{t}(A_{t}^{*})\right)
$$
and
$$\text{BayesRegret}\left(T,\,\pi^{\rm PS}\right) \leq \mathbb{E}\sum_{t=1}^{T}\left[U_{t}(A_{t})-f_{\theta}(A_{t})\right]+C\sum_{t=1}^{T}\mathbb{P}\left(f_{\theta}\left(A_{t}^{*}\right)>U_{t}(A_{t}^{*})\right).
$$
An important difference to take note of is that the Bayesian regret bound of $\pi^U$ depends on the specific upper confidence bound
sequence $U$ used by the UCB algorithm in question whereas the bound of $\pi^{\rm PS}$ applies simultaneously for {\it all}
upper confidence bound sequences.  This suggests that, while the Bayesian regret of a UCB algorithm depends critically on the 
specific choice of confidence sets, posterior sampling depends on the best possible choice of confidence sets.  This is 
a crucial advantage when there are complicated dependencies among actions, 
as designing and computing with appropriate confidence sets presents significant challenges. 
This difficulty is likely the main reason that posterior sampling significantly outperforms recently 
proposed UCB algorithms in the simulations presented in Section \ref{sec:simulation}. 

We have shown how upper confidence bounds characterize Bayesian regret bounds for posterior 
sampling.  We will leverage this concept in the next two sections.  Let us emphasize, though, that 
while our {\it analysis} of posterior sampling will make use of upper confidence bounds,
the actual {\it performance} of posterior sampling does not depend on upper confidence
bounds used in the analysis. 

\section{From UCB to Posterior Sampling Regret Bounds.}
\label{sec:UCB}

In this section we present Bayesian regret bounds for posterior sampling that can be
derived by combining our regret decomposition (\ref{eq: posterior sampling Risk Decomposition})
with results from prior work on UCB regret bounds.  Each UCB 
regret bound was established through a common procedure, which entailed 
specifying lower and upper confidence bounds $L_t:\mathcal{A}\mapsto \mathbb{R}$ and $U_t:\mathcal{A}\mapsto \mathbb{R}$ 
so that $L_t(a) \leq f_\theta(a) \leq U_t(a)$ with high probability for each $t$ and $a$,
and then providing an expression that dominates the sum $\sum_{1}^{T}(U_{t}-L_{t})(a_{t})$
for all sequences of actions $a_{1},..,a_{T}$.  As we will show, each such 
analysis together with our regret decomposition (\ref{eq: posterior sampling Risk Decomposition})
leads to a Bayesian regret bound for posterior sampling.

\subsection{Finitely Many Actions.} \label{subsec: finite}

We consider in this section a problem with $|\mathcal{A}| < \infty$ actions
and rewards satisfying $R_t \in [0,1]$ for all $t$ almost surely. We note, however,
that the results we discuss can be extended to cases where $R_t$ is not bounded
but where instead its distribution is ``light-tailed.''  It is also worth noting that
we make no further assumptions on the class of reward functions $\mathcal{F}$
or on the prior distribution over $\theta$.

In this setting, Algorithm \ref{alg:independentUCB}, which was proposed by \citet{auer2002finite},
is known to satisfy a problem-independent regret bound of order $\sqrt{|\mathcal{A}| T \log T}$. 
Under an additional assumption that action rewards are independent and take values in $\{0,1\}$,
an order $\sqrt{|\mathcal{A}| T \log T}$ regret bound for posterior sampling is also 
available \cite{agrawal2012further}. 

Here we provide a Bayesian regret bound that is
also of order $\sqrt{|\mathcal{A}| T \log T}$ but does not require that action rewards are
independent or binary.  Our analysis, like that of \citet{auer2002finite},
makes use of confidence sets that are Cartesian products of action-specific 
confidence intervals.  The regret decomposition 
(\ref{eq: posterior sampling Risk Decomposition}) lets us use 
such confidence sets to produce bounds for posterior sampling 
even when the algorithm itself may exploit dependencies among actions.

\begin{proposition}\label{prop:finite}
If $|\mathcal{A}|=K<\infty$ and $R_t \in [0,1]$, then for any $T\in \mathbb{N}$ 
\begin{equation}
\label{eq: high probability finite action bound}
\mathrm{BayesRegret}(T, \pi^{\mathrm{PS}}) \leq 2\min\{K, T\}+4\sqrt{K T(2+6\log(T))}.
\end{equation}
\end{proposition}
\proof{\textup{Proof.}}
Let $N_{t}(a)=\sum_{l=1}^{t}\mathbf{1}(A_t=a)$ denote the number of times $a$ is sampled over the first $t$ periods, and $\hat{\mu}_t(a)=N_{t}(a)^{-1}\sum_{l=1}^{t}\mathbf{1}(A_t=a)R_t$ denote the empirical average reward from these samples. Define upper and lower confidence bounds as follows:
\begin{eqnarray}\label{eq: finiteucb}
U_t(a) =\min\left\{\hat{\mu}_{t-1}(a) + \sqrt{\frac{2+6\log(T)}{N_{t-1}(a)}}, 1  \right\}  &\,& L_t(a) =\max\left\{\hat{\mu}_{t-1}(a) - \sqrt{\frac{2+6\log(T)}{N_{t-1}(a)}}, 0 \right\}. \end{eqnarray}
The next lemma, which is a consequence of analysis in \citet{abbasi2011improved}, shows these hold with high probability. Were actions sampled in an iid fashion, this lemma would follow immediately from the Hoeffding inequality. For more details, see Appendix \ref{sec: simpleconcentration}.  
\begin{lemma}\label{lem: finite case confidence} If $U_t(a)$ and $L_t(a)$ are defined as in \eqref{eq: finiteucb}, then 
$\mathbb{P}\left( \bigcup_{t=1}^{T} \{f_{\theta}(a) \notin [L_{t}(a), U_t(a)] \}  \right) \leq 1/T$.
\end{lemma}
First consider the case where $T\leq K$. Since $f_{\theta}(a)\in [0,1]$, $\mathrm{BayesRegret}(T, \pi^{\rm PS}) \leq T = \min\{K,T\}$.

Now, assume $T>K$.  Then,
\begin{eqnarray*}
\mathrm{BayesRegret}(T, \pi^{\mathrm{PS}}) &\leq& \mathbb{E}\left[  \sum_{t=1}^{T} (U_{t}-L_{t})(A_t) \right]+ T\mathbb{P}\left( \bigcup_{a\in\mathcal{A}} \bigcup_{t=1}^{T}  \left\{f_{\theta}(a) \notin [L_{t}(a), U_t(a)] \right\} \right) \\
&\leq& \mathbb{E}\left[ \sum_{t=1}^{T} (U_{t}-L_{t})(A_t) \right]+K.\\
\end{eqnarray*}

We now turn to bounding $\sum_{t=1}^{T} (U_{t}-L_{t})(A_t) $. Let $\mathcal{T}_a=\left\{t\leq T: A_t=a  \right\}$ denote the periods in which $a$ is selected. Then, $\sum_{t=1}^{T} (U_{t}-L_{t})(A_t) = \sum_{a\in\mathcal{A}} \sum_{t\in \mathcal{T}_a} (U_{t}-L_t)(a)$. We show,   
\begin{equation*}
\sum_{t\in \mathcal{T}_a} (U_{t}-L_t)(a) \leq 1+2\sqrt{2+6\log(T)}\sum_{t\in \mathcal{T}_a} (1+N_{t-1}(a))^{-1/2}= 1+2\sqrt{2+6\log(T)} \sum_{j=0}^{N_{T}(a)-1} (j+1)^{-1/2}
\end{equation*}
and 
\begin{equation*}
\sum_{j=0}^{N_{T}(a)-1} (j+1)^{-1/2}  \leq \intop_{x=0}^{N_{T}(a)} x^{-1/2}dx = 2\sqrt{N_{T}(a)}.
\end{equation*}

Summing over actions and applying the cauchy-shwartz inequality yields, 
\begin{eqnarray*}
\mathrm{BayesRegret}(T, \pi^{\rm PS}) \leq
2K+ 4\sqrt{2+6\log(T)}\sum_{a\in\mathcal{A}}\sqrt{ N_{T}(a)}&\overset{(a)}{\leq}& 2K+ 4\sqrt{(2+6\log(T))K \sum_{a}N_{T}(a)} \\
&=&2K+4\sqrt{KT(2+6\log(T))}\\
&\overset{(b)}{=}&2\min\{K,T\}+4\sqrt{KT(2+6\log(T))},
\end{eqnarray*}
where (a) follows from the cauchy-shwartz inequality and (b) follows from the assumption that $T> K$. \Halmos
\endproof

\subsection{Linear and Generalized Linear Models.}
\label{subsec: linear bandits}

We now consider function classes that represent linear and generalized linear models.  The bound of Proposition \ref{prop:finite}
applies so long as the number of actions is finite, but we will establish alternative bounds that depend on the dimension of the function 
class rather than the number of actions.  Such bounds accommodate problems with infinite action sets and can be much stronger than 
the bound of Proposition \ref{prop:finite} if there are many actions.

The Bayesian regret bounds we provide in this section derive from regret bounds of the UCB literature.  In Section \ref{sec:dimension},
we will establish a Bayesian regret bound that is as strong for the case of linear models and stronger for the case of generalized linear
models.  Since the results of Section \ref{sec:dimension} to a large extent supersede those we present here, we  
aim to be brief and avoid formal proofs in this section's discussion of the bounds and how they follow from results in the literature.

\subsubsection{Linear Models}

In the ``linear bandit'' problem studied by \cite{dani2008stochastic, rusmevichientong2010linearly, abbasi2011improved, abbasi2012online},
reward functions are parameterized by a vector $\theta\in\Theta\subset\mathbb{R}^d$,
and there is a known feature mapping $\phi:\mathcal{A} \mapsto \mathbb{R}^d$ such that $f_\theta(a) = \left\langle \phi(a), \theta \right\rangle$.
The following proposition establishes Bayesian regret bounds for such problems.  The proposition uses the term $\sigma$-sub-Gaussian
to describe any random variable $X$ that satisfies $\mathbb{E}\exp(\lambda X) \leq \exp(\lambda^2 \sigma^2 /2)$ for all $\lambda \in \mathbb{R}$.
\begin{proposition}
\label{prop:linear}
Fix positive constants $\sigma$, $c_1$, and $c_2$.  If $\Theta\subset\mathbb{R}^d$, $f_\theta(a)=\left\langle \phi(a), \theta \right\rangle$ for some
$\phi:\mathcal{A} \mapsto \mathbb{R}$, $\sup_{\rho \in \Theta} \|\rho\|_2 \leq c_1$, and
$\sup_{a \in \mathcal{A}} \|\phi(a)\|_2 \leq c_2$, and for each $t$, $R_t - f_{\theta}(A_t)$ conditioned
on $(H_t, A_t, \theta)$ is $\sigma$-sub-Gaussian, then
$${\rm BayesRegret}\left(T,\,\pi^{\rm PS}\right) = O(d \log T \sqrt{T}),$$
and
$${\rm BayesRegret}\left(T,\,\pi^{\rm PS}\right) = \tilde{O}\left(\mathbb{E} \sqrt{ \left\Vert  \theta \right\Vert_0 dT}  \right).$$
\end{proposition}
The second bound essentially replaces the dependence on the dimension $d$ with one on $\mathbb{E} \sqrt{\left\Vert  \theta \right\Vert_0 d}$.
The ``zero-norm'' $\left\Vert  \theta \right\Vert_0$ is the number of nonzero components, which can be much smaller than $d$
when the reward function admits a sparse representation.  Note that $\tilde{O}$ ignores logarithmic factors.
Both bounds follow from our regret decomposition 
(\ref{eq: posterior sampling Risk Decomposition}) together with
the analysis of \cite{abbasi2011improved}, in the case of the first bound, and the analysis of \cite{abbasi2012online},
in the case of the second bound.  We now provide a brief sketch of how these bounds can be derived.

If $f_{\theta}$ takes values in $[-C,\, C]$ then (\ref{eq: posterior sampling Risk Decomposition}) implies
\begin{equation}
\text{BayesRegret}\left(T,\,\pi^{\rm PS}\right) \leq \mathbb{E}\sum_{t=1}^{T}\left[U_{t}(A_{t})-L_{t}(A_{t})\right]+2C\sum_{t=1}^{T}\left[\mathbb{P}\left(f_{\theta}(A^{*}_{t})>U_{t}(A^{*}_{t})\right)+\mathbb{P}\left(f_{\theta}(A_{t})<L_{t}(A_{t})\right)\right].
\label{eq: bounded risk decomposition}
\end{equation}
The analyses of \cite{abbasi2011improved} and \cite{abbasi2012online} follow two steps that can be used to 
bound the right hand side of this equation.  In the first step, an ellipsoidal confidence set 
$\Theta_t:=\{\rho\in\mathbb{R}^d: \| \rho-\hat{\theta}_t \|_{V_t}\leq \sqrt{\beta_t}\}$
is constructed, where for some $\lambda\in\mathbb{R}$, $V_t:=\sum_{k=1}^{t} \phi(A_t)\phi(A_t)^{T}+\lambda I$ 
captures the amount of exploration carried out in each direction up to time $t$.
The upper and lower bounds induced by the ellipsoid are
$U_t(a):=\max \left\{C,  \max_{\rho\in\Theta_t}\left( \rho^{T}\phi(a) \right) \right\}$ and
$L_t(a):=\min \left\{-C,  \min_{\rho\in\Theta_t}\left( \rho^{T}\phi(a) \right) \right\}$.
If the sequence of confidence parameters $\beta_1,\ldots, \beta_T$ is selected so that
$\mathbb{P}(\theta \notin \Theta_t | H_t) \leq 1 / T$ then the second term of the regret decomposition
is less than $4C$.  For these confidence sets, the second step establishes a bound on $\sum_{1}^{T}(U_{t}-L_{t})(a_{t})$ 
that holds for any sequence of actions.  
The analyses presented on pages 7-8 of \cite{dani2008stochastic} and pages 14-15 of \cite{abbasi2011improved}
each implies such a bound of order $\sqrt{d \max_{t\leq T}\beta_tT\log (T/\lambda)}$.  Plugging in closed
form expressions for $\beta_t$ provided in these papers leads to the bounds of Proposition \ref{prop:linear}.

\subsubsection{Generalized Linear Models}

In a generalized linear model, the reward
function takes the form $f_{\theta}(a) := g\left(  \left\langle \phi(a), \theta\ \right\rangle\right)$
where the inverse link function $g$ is strictly increasing and continuously differentiable.
The analysis of \cite{filippi2010parametric} can be applied to establish a Bayesian regret 
bound for posterior sampling,
but with one caveat. The algorithm considered in \cite{filippi2010parametric} begins 
by selecting a sequence of actions $a_1,..,a_d$ with linearly independent 
feature vectors $\phi(a_1), \ldots, \phi(a_d)$.  Until now, we haven't even assumed such actions exist
or that they are guaranteed to be feasible over the first $d$ time periods. 
After this period of designed exploration, the algorithm selects at each time an action that maximizes
an upper confidence bound.  What we will establish using the results from \cite{filippi2010parametric}
is a bound on a similarly modified version of posterior sampling, in which the first $d$
actions taken are $a_1, \ldots, a_d$, while subsequent actions are selected by posterior sampling.
Note that the posterior distribution employed at time $d+1$ is conditioned on observations made
over the first $d$ time periods.  We denote this modified posterior sampling
algorithm by $\pi^{\rm IPS}_{a_1,\ldots,a_d}$.  It is worth mentioning here that in Section 
\ref{sec:dimension} we present a result with a stronger bound that applies to the standard version
of posterior sampling, which does not include a designed exploration period.
\begin{proposition}
\label{prop:generalizedLinear}
Fix positive constants $c_1$, $c_2$, $C$, and $\lambda$.
If $\Theta\subset\mathbb{R}^d$, $f_\theta(a)=g(\left\langle \phi(a), \theta \right\rangle)$ for some strictly
increasing continuously differentiable function $g:\mathbb{R}\mapsto \mathbb{R}$, 
 $\sup_{\rho \in \Theta} \|\rho\|_2 \leq c_1$, $\sup_{a \in \mathcal{A}} \|\phi(a)\|_2 \leq c_2$,
 $\mathcal{A}_t = \mathcal{A}$ for all $t$,
$\sum_{i=1}^{d} \phi(a_i) \phi(a_i)^T \succeq \lambda I$ for some $a_1,\ldots,a_d \in \mathcal{A}$, and $R_t \in [0,C]$
for all $t$, then
$${\rm BayesRegret}\left(T,\,\pi^{\rm IPS}_{a_1,\ldots,a_d}\right) = O(r d \log^{3/2} T \sqrt{T}),$$
where $r=\sup_{\rho,a}g'(\langle \phi(a), \rho\rangle) / \inf_{\rho,a}g'(\langle \phi(a),\rho\rangle)$.
\end{proposition}
Like the analyses of \cite{abbasi2011improved} and \cite{abbasi2012online}, which apply to linear models,
the analysis of \cite{filippi2010parametric} follows two steps that together bound both terms of our regret
decomposition (\ref{eq: bounded risk decomposition}).  First, an ellipsoidal confidence set 
$\Theta_t$ is constructed, centered around a quasi-maximum likelihood estimator.
This confidence set is designed to contain $\theta$ with high probability. 
Given confidence bounds  
$U_t(a):=\max\{C,  \max_{\rho\in\Theta_t}g(\langle\phi(a), \rho \rangle) \}$ and
$L_t(a):=\min\{0,  \min_{\rho\in\Theta_t}g(\langle \phi(a), \rho \rangle)\}$,
a worst case bound on $\sum_{1}^{T}(U_{t}-L_{t})(a_{t})$ is established.
The bound is similar to those established for the linear case, but there is
an added dependence on the the slope of $g$.

\subsection{Gaussian Processes.}\label{subsec:GP}

In this section we consider the case where the
reward function $f_\theta $ is sampled from a Gaussian process.
That is, the stochastic process 
$\left( f_\theta(a):  a\in \mathcal{A} \right)$ is such that for any
$a_1,..,a_k \in \mathcal{A}$ the collection $f_{\theta}(a_1),..,f_{\theta}(a_k)$ 
follows a multivariate Gaussain distribution. \citet{srinivas2012information} 
study a UCB algorithm designed for such problems and provide general regret bounds. 
Again, through the regret decomposition (\ref{eq: posterior sampling Risk Decomposition})
 their analysis provides a Bayesian regret bound for posterior sampling. 

For simplicity, we focus our discussion on the case where $\mathcal{A}$ is finite,
so that $\left( f_\theta(a):  a\in \mathcal{A} \right)$ follows a multivariate Gaussian distribution.  
As shown by \citet{srinivas2012information}, the results extend to infinite action sets through a 
discretization argument as long as certain smoothness conditions are satisfied. 

When confidence bounds hold, a UCB algorithm incurs high regret from sampling an action 
only when the confidence bound at that action is loose. In that case, one would expect
the algorithm to learn a lot about $f_\theta$ based on the observed reward. This suggests
the algorithm's cumulative regret may be bounded in an appropriate sense by the total 
amount it is expected to learn. Leveraging the structure of the Gaussian distribution, 
\citet{srinivas2012information} formalize this idea. 
They bound the regret of their UCB algorithm in terms of the maximum amount that 
{\it any} algorithm could learn about $f_\theta$. They use an information theoretic 
measure of learning: the information gain. This is defined to be the 
difference between the entropy of the prior distribution of 
$\left( f_\theta(a):  a\in \mathcal{A} \right)$ and
the entropy of the posterior. The maximum possible information 
gain is denoted $\gamma_T$, where the maximum is
taken over all sequences $a_1,..,a_T$.%
\footnote{An important property of the Gaussian distribution is that the information
gain does not depend on the observed rewards. This is because the
posterior covariance of a multivariate Gaussian is a deterministic
function of the points that were sampled. For this reason, this maximum
is well defined. %
}
Their analysis also supports the following result on posterior sampling. 
\begin{proposition}
If $\mathcal{A}$ is finite, $\left( f_\theta(a):  a\in \mathcal{A} \right)$ follows a multivariate 
Gaussian distribution with marginal variances bounded by $1$,
$R_t-f_{\theta}(A_t)$ is independent of $(H_t, \theta, A_t)$, and
$\{R_t-f_{\theta}(A_t) | t \in \mathbb{N}\}$ is an iid sequence of zero-mean Gaussian random
variables with variance $\sigma^{2}$, then 
$${\rm BayesRegret}\left(T,\,\pi^{\rm PS}\right)\leq 1+2\sqrt{T \gamma_T \ln\left(1+\sigma^{-2}\right)^{-1} \ln\left(\frac{(T^2+1) \left| \mathcal{A}\right|} {\sqrt{2\pi}} \right)}$$
for all $T \in \mathbb{N}$.
\end{proposition}

\citet{srinivas2012information} also provide bounds on $\gamma_{T}$ 
for kernels commonly used in Gaussian process regression, including the linear kernel,
radial basis kernel, and Mat\'{e}rn kernel. Combined with the above
proposition, this yields explicit Bayesian regret bounds in these cases. 

We will briefly comment on their analysis and how it provides a bound for
posterior sampling. First, note that the posterior distribution is Gaussian,
which suggests an upper confidence bound of the form 
$U_{t}(a):=\mu_{t-1}(a)+\sqrt{\beta_{t}}\sigma_{t-1}(a)$, where $\mu_{t-1}(a)$ is the 
posterior mean, $\sigma_{t-1}(a)$ is the posterior standard deviation of $f_{\theta}(a)$, 
and $\beta_t$ is a confidence parameter.
We can provide a Bayesian regret bound by bounding both terms of 
(\ref{eq: posterior sampling Risk Decomposition}).  The next lemma, which follows,  bounds the second term. 
Some new analysis is required since the Gaussian distribution is unbounded, and we study Bayesian regret whereas \citet{srinivas2012information} 
bound regret with high probability under the prior. 

\begin{lemma}
\label{lem: gaussian tail bound}
If $U_{t}(a):=\mu_{t-1}(a)+\sqrt{\beta_{t}}\sigma_{t-1}(a)$ and 
$\beta_t:=2\ln\left( \frac{\left(t^{2}+1\right) \left| \mathcal{A}\right|}{\sqrt{2\pi}} \right)$ then for all $T \in \mathbb{N}$
$\mathbb{E}\sum_{t=1}^{T}\left[f_{\theta}\left(A_{t}^{*}\right)-U_{t}(A_{t}^{*})\right]\leq1.$
\end{lemma}
\proof{\textup{Proof.}} \noindent  First, if $X\sim N(\mu, \sigma^2)$ then if $\mu \leq 0$, 
$
\mathbb{E} \left[X {\bf 1}\left\{X>0 \right\} \right]=\intop_{0}^{\infty}\frac{x}{\sigma\sqrt{2\pi}}\exp\left\{ \frac{-(x-\mu)^{2}}{2\sigma^{2}}\right\} dx=\frac{\sigma}{\sqrt{2\pi}}\exp\left\{ \frac{-\mu^{2}}{2\sigma^{2}}\right\}. 
$

\noindent Then since the posterior distribution of $f_{\theta}(a)-U_{t}(a)$ is normal with mean $-\sqrt{\beta_{t}}\sigma_{t-1}(a)$ and variance $\sigma^2_{t-1}(a)$
\begin{equation} \label{eq:Gaussian Tail Sum}
\mathbb{E}\left[\mathbf{1}\left\{ f_{\theta}(a)-U_{t}(a) \geq 0 \right\} \left[f_{\theta}(a)-U_{t}(a)\right] |H_t \right] 
=\frac{\sigma_{t-1}(a)}{\sqrt{2\pi}}\exp\left\{ \frac{-\beta_t}{2}\right\} 
=\frac{\sigma_{t-1}(a)}{\left(t^{2}+1\right)\left|\mathcal{A}\right|} \leq \frac{1}{\left(t^{2}+1\right)\left|\mathcal{A}\right|}.
\end{equation}
\noindent  The final inequality above follows from the assumption that $\sigma_0(a)\leq1$. The claim follows from (\ref{eq:Gaussian Tail Sum}) since
$$
\mathbb{E}\sum_{t=1}^{T}\left[f_{\theta}(A_{t}^{*})-U_{t}(A_{t}^{*})\right] \leq
\sum_{t=1}^{\infty}\sum_{a\in\mathcal{A}}\mathbb{E}\left[\mathbf{1}\left\{ f_{\theta}(a)-U_{t}(a)\geq 0\right\} \left[f_{\theta}(a)-U_{t}(a)\right]\right]
\leq \sum_{t=1}^{\infty} \frac{1}{\left(t^{2}+1\right)} \leq 1. \Halmos
$$
\endproof

\noindent Now, consider the first term of (\ref{eq: posterior sampling Risk Decomposition}), which is:
$$\mathbb{E}\sum_{t=1}^{T}(U_{t}-f_{\theta})(A_{t})=\mathbb{E}\sum_{t=1}^{T}(U_{t}-\mu_{t-1})(A_{t})=\mathbb{E}\sum_{t=1}^{T}\sqrt{\beta_{t}}\sigma_{t-1}(A_t)
\leq \mathbb{E}\sqrt{T \max_{t\leq T}\beta_t}\sqrt{\sum_{t=1}^{T}{\sigma^2_{t-1}(A_t)}}.$$
Here the second equality follows by the tower property of conditional expectation, 
and the final step follows from the Cauchy-Schwartz inequality. Therefore, to establish a Bayesian regret bound it is sufficient to provide a bound
on the sum of posterior variances $\sum_{t=1}^{T}{\sigma^2_{t-1}(a_t)}$ that 
holds for any $a_1,..,a_T$.  Under the assumption that $\sigma_0(a)\leq1$, 
the proof of Lemma 5.4 of \citet{srinivas2012information} shows that
$\sigma^2_{t-1}(a_t) \leq \alpha^{-1} \log \left(1+\sigma^{-2} \sigma_{t-1}^{2}(a_t) \right)$, 
where $\alpha=\left(1+\sigma^{-2}\right)$. 
At the same time, Lemma 5.3 of \citet{srinivas2012information} shows
the information gain from selecting $a_1,...a_T$ is equal to 
$\frac{1}{2}\sum_{t=1}^{T}\log \left(1+\sigma^{-2} \sigma_{t-1}^{2}(a_t)     \right)$. 
This shows that for any actions $a_1,..,a_T$ the the sum of posterior variances 
$\sum_{t=1}^{T}{\sigma^2_{t-1}(a_t)}$ can be bounded
in terms of the information gain from selecting $a_1,..,a_T$. 
Therefore $\sum_{t=1}^{T}{\sigma^2_{t-1}(A_t)}$ can be bounded
in terms of the {\it largest possible}  information gain $\gamma_T$. 

\section{Bounds for General Function Classes.}
\label{sec:dimension}

The previous section treated models in which the relationship among
action rewards takes a simple and tractable form. Indeed, nearly all of the multi-armed bandit
literature focuses on such problems.  Posterior sampling can be applied to
a much broader class of models.  As such, more general results that hold 
beyond restrictive cases are of particular interest.  In this section, we provide a Bayesian regret
bound that applies when the reward function lies in a known, but otherwise
arbitrary class of uniformly bounded real-valued functions $\mathcal{F}$.
Our analysis of this abstract framework yields a more general result that applies beyond
the scope of specific problems that have been studied in the literature, and also
identifies factors that unify more specialized prior results.  Further, our
more general result when specialized to linear models recovers the strongest known
Bayesian regret bound and in the case of generalized linear models yields a bound
stronger than that established in prior literature.

If $\mathcal{F}$ is not appropriately restricted, it is impossible to guarantee any reasonably attractive
level of Bayesian regret.  
For example, in a case where $\mathcal{A} = [0,1]$, $f_\theta(a) = {\bf 1}(\theta = a)$, 
$\mathcal{F} = \{f_\theta | \theta \in [0,1]\}$, and $\theta$ is uniformly distributed over $[0,1]$,
it is easy to see that the Bayesian regret of {\it any} algorithm over $T$ periods is $T$, 
which is no different from the worst level of performance an agent can experience. 

This example highlights the fact that Bayesian regret bounds must depend on 
the function class $\mathcal{F}$.  The bound we develop in this section
depends on $\mathcal{F}$ through two measures of complexity.  
The first is the Kolmogorov dimension, which measures the growth 
rate of the covering numbers of $\mathcal{F}$
and is closely related to measures of complexity that are common in the supervised learning
literature. It roughly captures the sensitivity
of $\mathcal{F}$ to statistical over-fitting. The second measure is a new notion
we introduce, which we refer to as the eluder dimension.  This captures how
effectively the value of unobserved actions can be inferred
from observed samples. We highlight in Section \ref{subsec: vc}
why notions of dimension common to the supervised learning literature are
insufficient for our purposes. 


Though the results of this section are very general, they do not apply to the entire range of 
problems represented by the formulation we introduced in Section \ref{sec:formulation}.
In particular, throughout the scope of this section, we fix constants $C > 0$ and $\sigma > 0$ 
and impose two simplifying assumptions.  The first concerns boundedness of reward
functions.
\begin{assumption}
For all $f \in \mathcal{F}$ and $a \in \mathcal{A}$, $f(a) \in [0,C]$.
\end{assumption}
\noindent Our second assumption ensures that observation noise is light-tailed.
Recall that we say a random variable $x$ is $\sigma$-sub-Gaussian if 
$\mathbb{E}[\exp(\lambda x)]\leq\exp(\lambda^2 \sigma^2 /2)$ almost surely for all $\lambda$.
\begin{assumption}
For all $t \in \mathbb{N}$, $R_t-f_{\theta}(A_t)$ conditioned on $(H_t, \theta, A_t)$ 
is $\sigma$-sub-Gaussian.
\end{assumption}
\noindent  It is worth noting that the Bayesian regret bounds we provide are distribution independent, in the sense that we show $\mathrm{BayesRegret}(T, \pi^\mathrm{PS})$ is bounded by an expression that does not depend on $\mathbb{P}(\theta \in \cdot)$. 

Our analysis in some ways parallels those found in the literature on UCB algorithms.  
In the next section we provide a method for constructing a set $\mathcal{F}_{t}\subset\mathcal{F}$ of functions
that are statistically plausible at time $t$. Let $w_{\mathcal{F}}(a):=\sup_{\overline{f}\in\mathcal{F}}\overline{f}(a)-\inf_{\underline{f}\in\mathcal{F}}\underline{f}(a)$
denote the width of $\mathcal{F}$ at $a$. Based on these confidence
sets, and using the regret decomposition (\ref{eq: posterior sampling Risk Decomposition}),
one can bound Bayesian regret in terms of $\sum_{1}^{T}w_{\mathcal{F}_t}(A_t)$.
In Section \ref{subsec: Bayesian regret}, we establish a bound on this sum in terms of
the Kolmogorov and eluder dimensions of $\mathcal{F}$.

\subsection{Confidence Bounds.}
\label{subsec: confidence bounds}

The construction of tight confidence sets for specific classes of functions presents technical challenges.  
Even for the relatively simple case of linear bandit problems,  
significant analysis is required. It is therefore perhaps surprising that, as we show in this section, 
one can construct strong confidence sets for an arbitrary class of functions 
without much additional sophistication.  While the focus of our work is
on providing a Bayesian regret bound for posterior sampling, the techniques we introduce
for constructing confidence sets may find broader use.


The confidence sets constructed here are centered around least squares
estimates $\hat{f}_{t}^{LS}\in\arg\min_{f\in\mathcal{F}}L_{2,t}(f)$
where $L_{2,t}(f)=\sum_{1}^{t-1} (f(A_{t})-R_{t})^{2}$ 
is the cumulative squared prediction error.\footnote{The results 
can be extended to the case where the infimum of
$L_{2,t}(f)$ is unattainable by selecting a function with squared prediction error 
sufficiently close to the infimum. %
} The sets take the form $\mathcal{F}_{t}:= \{ f\in\mathcal{F}: \|f-\hat{f}_{t}^{LS}\|_{2,E_{t}}\leq\sqrt{\beta_{t}}\} $
where $\beta_{t}$ is an appropriately chosen confidence parameter,
and the empirical 2-norm $\left\Vert \cdot\right\Vert _{2,E_{t}}$
is defined by $\left\Vert g\right\Vert _{2,E_{t}}^{2}=\sum_{1}^{t-1}g^2(A_{k})$. 
Hence $\left\Vert f-f_{\theta}\right\Vert _{2, E_{t}}^{2}$ measures the cumulative discrepancy
between the previous predictions of $f$ and $f_\theta$.


The following lemma is the key to constructing strong confidence sets
$\left(\mathcal{F}_{t}:t\in\mathbb{N}\right)$. For an arbitrary function
$f$, it bounds the squared error of $f$ from below in terms of the
empirical loss of the true function $f_{\theta}$ and the aggregate
empirical discrepancy $\left\Vert f-f_{\theta}\right\Vert _{2,E_{t}}^{2}$
between $f$ and $f_{\theta}$. It establishes that for any function
$f$, with high probability, the random process $(L_{2,t}(f) : t\in\mathbb{N})$
never falls below the process $(L_{2,t}(f_{\theta})+\frac{1}{2}\| f-f_{\theta}\|_{2,E_{t}}^{2}:\, t\in\mathbb{N})$
by more than a fixed constant. A proof of the lemma is provided in the appendix.
\begin{lemma}
\label{lem: least squares bound}
For any $\delta > 0$ and $f:\mathcal{A} \mapsto \mathbb{R}$, 
with probability at least $1-\delta$,
\[
L_{2,t}(f)\geq L_{2,t}(f_{\theta})+\frac{1}{2}\left\Vert f-f_{\theta}\right\Vert _{2,E_{t}}^{2}-4\sigma^{2}\log\left(1/\delta\right)
\]
simultaneously for all $t \in \mathbb{N}$.
\end{lemma}

By Lemma \ref{lem: least squares bound}, with high probability, $f$ can 
enjoy lower squared error than $f_{\theta}$ only if its empirical deviation
$\left\Vert f-f_{\theta}\right\Vert _{2,E_{t}}^{2}$ from $f_{\theta}$
is less than $8\sigma^{2}\log(1/\delta)$. Through a union bound,
this property holds uniformly for all functions in a finite subset
of $\mathcal{F}$. Using this fact and a discretization
argument, together with the observation that $L_{2,t}(\hat{f}_{t}^{LS})\leq L_{2,t}(f_{\theta})$,
we can establish the following result, which is proved in the appendix.
Let $N(\mathcal{F},\,\alpha,\,\left\Vert \cdot\right\Vert _{\infty})$
denote the $\alpha$-covering number of $\mathcal{F}$ in the sup-norm
$\| \cdot\|_{\infty}$, and let
\begin{equation}
\beta^{*}_{t} \left( \mathcal{F}, \delta, \alpha\right) :=
8\sigma^{2}\log\left(N(\mathcal{F},\,\alpha,\,\left\Vert \cdot\right\Vert _{\infty})/\delta\right)+2\alpha t\left(8C+\sqrt{8\sigma^{2}\ln(4t^{2}/\delta)}\right). \label{eq:Confidence Parameter}
\end{equation}
\begin{proposition}
\label{prop: least squares bound}
For all $\delta > 0$ and $\alpha > 0$, if 
$$\mathcal{F}_t=\left\{ f\in\mathcal{F}:\,\left\Vert f-\hat{f}_t^{LS}\right\Vert _{2,E_t}
\leq\sqrt{\beta^{*}_t \left( \mathcal{F}, \delta, \alpha\right)}    \right\}$$
for all $t \in \mathbb{N}$, then
\[
\mathbb{P}\left(f_{\theta}\in\bigcap_{t=1}^{\infty}\mathcal{F}_{t}\right)\geq1-2\delta.
\]
\end{proposition}

While the expression (\ref{eq:Confidence Parameter}) defining the confidence parameter is complicated, 
it can be bounded by simple expressions in important cases.  We provide three examples.
\begin{example}
{\bf Finite function classes:}
When $\mathcal{F}$ is finite,
$\beta^*_t(\mathcal{F}, \delta, 0) = 8\sigma^2 \log(\left|\mathcal{F}\right|/\delta)$.
\end{example}

\begin{example}
\label{ex: beta linear models}
{\bf Linear Models:}
Consider the case of a $d$-dimensional linear model $f_{\rho}(a):=\left\langle \phi(a),\,\rho\right\rangle$.
Fix $\gamma = \sup_{a \in \mathcal{A}} \| \phi(a)\|$ and $s = \sup_{\rho \in \Theta} \|\rho\|$.  Hence, for all
$\rho_1, \rho_2 \in \mathcal{F}$, we have $\| f_{\rho_1}- f_{\rho_2}  \|_{\infty}\leq\gamma\|  \rho_1- \rho_2  \|$.
An $\alpha$-covering of $\mathcal{F}$ can therefore be attained through an $(\alpha / \gamma)$-covering of $\Theta\subset\mathbb{R}^{d}$. 
Such a covering requires $O((1 / \alpha)^d)$ elements, and it follows that, 
$\log N(\mathcal{F},\,\alpha,\,\left\Vert \cdot\right\Vert _{\infty})=O(d \log (1/\alpha))$. If $\alpha$ is chosen 
to be $1/t^2$, the second term in ($\ref{eq:Confidence Parameter}$) tends to zero, and therefore, 
$\beta_t^*(\mathcal{F}, \delta, 1/t^2) =O(d \log(t/\delta))$.
\end{example}

\begin{example}
\label{ex: beta generalized linear models}
{\bf Generalized Linear Models:}
Consider the case of a $d$-dimensional generalized linear model $f_{\theta}(a):=g\left(\left\langle \phi(a),\,\theta\right\rangle\right)$ where
$g$ is an increasing Lipschitz continuous function. Fix $g$, $\gamma = \sup_{a \in \mathcal{A}} \| \phi(a)\|$, 
and $s = \sup_{\rho \in \Theta} \|\rho\|$.  Then, the previous argument 
shows $\log N(\mathcal{F},\,\alpha,\,\left\Vert \cdot\right\Vert _{\infty})=O(d \log (1/\alpha))$.  
Again, choosing $\alpha=1/t^2$ yields a confidence parameter 
$\beta_t^*(\mathcal{F}, \delta, 1/t^2)=O(d \log(t/\delta))$.
\end{example}

The confidence parameter $\beta^*_t(\mathcal{F}, 1/t^2, 1/t^2)$ is closely related to the following concept.
\begin{definition}
\label{def: Kolmogorov}
The {\it Kolmogorov dimension} of a function class $\mathcal{F}$ is given by
\[
{\rm dim}_{K}(\mathcal{F})=\underset{\alpha\downarrow0}{\lim\sup}\frac{\log N(\mathcal{F},\,\alpha,\,\left\Vert \cdot\right\Vert _{\infty})}{\log(1/\alpha)}.
\]
\end{definition}
\noindent In particular, we have the following result.
\begin{proposition}
\label{prop: kolmogorov}
For any fixed class of functions $\mathcal{F}$,
$$\beta^*_t \left( \mathcal{F},  1/t^2,  1/t^2\right) =16(1+o(1)+{\rm dim}_{K}(\mathcal{F}))\log t .$$
\end{proposition}
\proof{\textup{Proof.}}  By definition
\[
\beta^*_t \left( \mathcal{F}, 1 / t^2,  1/ t^2\right)=
8\sigma^{2}\left[\frac{\log\left(N\left(\mathcal{F},\,1/t^2,\,\left\Vert \cdot\right\Vert _{\infty}\right)\right)}{\log\left(t^2\right)}+1\right]\log\left(t^2\right)+2\frac{t}{t^2} 
\left(8C+\sqrt{8\sigma^{2}\ln(4 t^2\delta)}\right)
\]
The result follows from the fact that 
$\underset{t\rightarrow\infty}{\lim\sup} \log\left(N\left(\mathcal{F},\,1/t^2,\,\left\Vert \cdot\right\Vert _{\infty}\right)\right)/\log\left(t^2\right)
={\rm dim}_{K}(\mathcal{F})$. \Halmos
\endproof

\subsection{Bayesian Regret Bounds.}
\label{subsec: Bayesian regret}

In this section we introduce a new notion of complexity -- the {\it eluder dimension} --
and then use it to develop a Bayesian regret bound.  First, we note that,
using the regret decomposition (\ref{eq: posterior sampling Risk Decomposition})
and the confidence sets $\left(\mathcal{F}_{t}:t\in\mathbb{N}\right)$ constructed in
the previous section, we can bound the Bayesian regret of
posterior sampling in terms confidence interval widths $w_{\mathcal{F}}(a):= \sup_{f \in \mathcal{F}} f(a) - \inf_{f \in \mathcal{F}} f(a)$.
In particular, the following lemma follows from our regret decomposition (\ref{eq: posterior sampling Risk Decomposition}). 
\begin{lemma}
\label{lem: risk under condition 1}
For all $T \in \mathbb{N}$, if $\inf_{\rho \in \mathcal{F}_\tau} f_\rho(a) \leq f_\theta(a) \leq \sup_{\rho \in \mathcal{F}_\tau} f_\rho(a)$ 
for all $\tau \in \mathbb{N}$ and $a \in \mathcal{A}$ with probability at least $1 - 1/T$ 
then
\[
{\rm BayesRegret}(T, \pi^{\rm PS}) \leq C+\mathbb{E}\sum_{t=1}^{T}w_{\mathcal{F}_t}(A_{t}).
\]
\end{lemma}
\noindent We can use the confidence sets constructed in the previous section
to guarantee that the conditions of this lemma hold.
In particular, choosing $\delta=1/2T$
in (\ref{eq:Confidence Parameter}) guarantees that $f_{\theta}\in\bigcap_{t=1}^{\infty}\mathcal{F}_{t}$
with probability at least $1-1/T$.

Our remaining task is to provide a worst case bound on the sum $\sum_{1}^{T}w_{\mathcal{F}_t}(A_{t})$.
First consider the case of a linearly parameterized model 
where $f_{\rho}(a):=\left\langle \phi(a),\,\rho\right\rangle $ 
for each $\rho\in\Theta\subset\mathbb{R}^d$. Then, it can be shown that our
confidence set takes the form 
$\mathcal{F}_t:=\left\{f_{\rho} : \rho\in\Theta_t\right\}$
where $\Theta_t\subset\mathbb{R}^d$ is an ellipsoid. When an action
$A_t$ is sampled, the ellipsoid shrinks in the direction $\phi(A_t)$. Here the 
explicit geometric structure of the confidence set implies that the width $w_{\mathcal{F}_t}$
shrinks not only at $A_t$ but also at any other action whose feature vector is  
not orthogonal to $\phi(A_t)$. Some linear algebra leads to a worst case bound 
on $\sum_{1}^{T}w_{\mathcal{F}_t}(A_{t})$.
For a general class of functions, the situation is much subtler, and we need to measure
the way in which the width at each action can be reduced by sampling other actions. 
To do this, we introduce the following notion of dependence. 

\begin{definition}
\label{def: independence}
An action $a \in \mathcal{A}$ is $\epsilon$-{\it dependent} on actions $\{a_{1},...,a_{n}\} \subseteq \mathcal{A}$ with respect
to $\mathcal{F}$ if any pair of functions $f, \tilde{f} \in \mathcal{F}$ satisfying
$\sqrt{\sum_{i=1}^{n} (f(a_i)-\tilde{f}(a_i))^{2}} \leq\epsilon$
also satisfies $f(a)-\tilde{f}(a) \leq \epsilon$.
Further, $a$ is $\epsilon$-independent of $\left\{a_{1},..,a_{n}\right\}$ with respect to $\mathcal{F}$
if $a$ is not $\epsilon$-dependent on $\left\{ a_{1},..,a_{n}\right\} $. 
\end{definition}

Intuitively, an action $a$ is independent of $\{a_{1},...,a_{n}\}$ if two
functions that make similar predictions  at $\{a_{1},...,a_{n}\}$ can nevertheless differ significantly 
in their predictions at $a$. The above definition measures the ``similarity'' of predictions at $\epsilon$-scale, 
and measures whether two functions make similar predictions  at $\{a_{1},...,a_{n}\}$ based on the 
cumulative discrepancy $\sqrt{\sum_{i=1}^{n} (f(a_i)-\tilde{f}(a_i))^{2}}$.
This measure of dependence suggests using the following notion of dimension.  In this definition, we
imagine that the sequence of elements in $\mathcal{A}$ is chosen by an {\it eluder} who hopes to show the agent poorly understood actions for as long as possible. 

\begin{definition}
\label{def: dimension}
The  $\epsilon$-{\it eluder dimension} ${\rm dim}_E(\mathcal{F}, \epsilon)$ is 
the length $d$ of the longest sequence of elements in $\mathcal{A}$ 
such that, for some $\epsilon' \geq \epsilon$, every element is $\epsilon'$-independent 
of its predecessors. 
\end{definition}

Recall that a vector space has 
dimension $d$ if and only if $d$ is the length of the longest sequence of elements
such that each element is linearly independent or equivalently, $0$-independent 
of its predecessors. Definition \ref{def: dimension} 
replaces the requirement of linear independence with $\epsilon$-independence. 
This extension is advantageous as it captures both nonlinear dependence and approximate dependence. 
The following result uses our new notion of dimension to bound the number of times the
width of the confidence interval for a selected action $A_t$ can exceed a threshold. 

\begin{proposition}
\label{prop: num_times_threshold_exceeded}
If $(\beta_t \geq 0 | t \in \mathbb{N})$ is a nondecreasing sequence and 
$\mathcal{F}_{t}:= \{f \in \mathcal{F}:\,\| f-\hat{f}_{t}^{LS}\|_{2,E_{t}}\leq\sqrt{\beta_{t}}\}$
then
$$\sum_{t=1}^T {\bf 1}(w_{\mathcal{F}_t}(A_t) > \epsilon) \leq \left(\frac{4 \beta_T}{\epsilon^2} + 1\right) {\rm dim}_E(\mathcal{F}, \epsilon)$$
for all $T \in \mathbb{N}$ and $\epsilon > 0$.
\end{proposition}
\proof{\textup{Proof.}}  We begin by showing that if $w_{t}(A_{t})>\epsilon$ then $A_t$ is $\epsilon$-dependent 
on fewer than $4\beta_T / \epsilon^2$ disjoint subsequences of $(A_{1},..,A_{t-1})$, for $T > t$.
To see this, note that if $w_{\mathcal{F}_t}(A_{t})>\epsilon$ there are $\bar{f},\,\underline{f}\,\in\mathcal{F}_{t}$
such that $\overline{f}(A_t) - \underline{f}(A_t) > \epsilon$.  By definition, since $\overline{f}(A_t) - \underline{f}(A_t) > \epsilon$, 
if  $A_{t}$ is $\epsilon$-dependent on a subsequence $(A_{i_{1}},..,\, A_{i_{k}})$ of $(A_{1},..,A_{t-1})$
then $\sum_{j=1}^{k}(\overline{f}(A_{i_{j}})-\underline{f}(A_{i_{j}}))^{2}>\epsilon^{2}$.  It follows that, if 
$A_{t}$ is $\epsilon$-dependent on $K$ disjoint subsequences of 
$(A_{1},..,A_{t-1})$ then $\|\overline{f}-\underline{f}\|_{2,E_{t}}^{2}>K\epsilon^{2}$.
By the triangle inequality, we have
$$\left\Vert \overline{f}-\underline{f}\right\Vert _{2,E_{t}}\leq\left\Vert \overline{f}-\hat{f}^{LS}_{t}\right\Vert _{2,E_{t}}+\left\Vert \underline{f}-\hat{f}^{LS}_{t}\right\Vert _{2,E_{t}}\leq2\sqrt{\beta_{t}} \leq2\sqrt{\beta_T}.$$
and it follows that $K< 4\beta_T / \epsilon^2$. 

Next, we show that in any action sequence $(a_1,..,a_\tau)$, there is some element $a_j$ that 
is $\epsilon$-dependent on at least $\tau/d -1$ disjoint subsequences of $\left(a_{1},..,a_{j-1}\right)$, where $d := {\rm dim}_E(\mathcal{F}, \epsilon)$. 
To show this, for an integer $K$ satisfying $Kd+1 \leq\tau\leq Kd+d$, we will construct $K$ disjoint subsequences
$B_1, \ldots, B_K$. First let $B_{i}=\left(a_{i}\right)$ for $i=1,..,K$. 
If $a_{K+1}$ is $\epsilon$-dependent on each subsequence $B_{1},..,B_{K}$,
our claim is established. Otherwise, select a subsequence $B_{i}$ such that $a_{K+1}$ is $\epsilon$-independent 
and append $a_{K+1}$ to $B_i$.  Repeat this process for elements with indices $j > K+1$ until $a_{j}$
is $\epsilon$-dependent on each subsequence or $j = \tau$.  In the latter scenario 
$\sum |B_i| \geq Kd$, and since each element of a subsequence $B_i$ is 
$\epsilon$-independent of its predecessors, $|B_i| = d$.  In this case, $a_\tau$ must be $\epsilon$--dependent
on each subsequence.

Now consider taking $(a_1,..,a_\tau)$ to be the subsequence $(A_{t_1}, \ldots, A_{t_\tau})$ of $(A_1,\ldots,A_T)$ consisting 
of elements $A_t$ for which $w_{\mathcal{F}_t}(A_{t})>\epsilon$.  As we have established, each $A_{t_j}$ is $\epsilon$-dependent 
on fewer than $4\beta_T / \epsilon^2$ disjoint subsequences of $(A_{1},..,A_{t_j-1})$.  It follows that each $a_j$
is $\epsilon$-dependent  on fewer than $4\beta_T / \epsilon^2$ disjoint subsequences of $(a_1,..,a_{j-1})$.
Combining this with the fact we have established that there is some $a_j$ that 
is $\epsilon$-dependent on at least $\tau/d -1$ disjoint subsequences of $\left(a_{1},..,a_{j-1}\right)$, 
we have $\tau/d -1 \leq 4\beta_T / \epsilon^2$.  It follows that $\tau \leq (4\beta_T / \epsilon^2 + 1) d$, which  is our
desired result.
\Halmos

Using Proposition \ref{prop: num_times_threshold_exceeded}, one can bound the sum 
$\sum_{t=1}^T w_{\mathcal{F}_t}(A_t)$, as established by the following lemma.
\begin{lemma}\label{lem: sum of widths}
If $(\beta_t \geq 0 | t \in \mathbb{N})$ is a nondecreasing sequence and 
$\mathcal{F}_t := \{f \in \mathcal{F}:\,\| f-\hat{f}_{t}^{LS}\|_{2,E_{t}}\leq\sqrt{\beta_{t}}\}$
then 

\[
\sum_{t=1}^{T}w_{\mathcal{F}_t}(A_t)\leq 1+\dim_{E}\left(\mathcal{F}, T^{-1} \right)C
+4\sqrt{\dim_{E}\left(\mathcal{F}, T^{-1} \right) \beta_{T}T}
\]
for all $T \in \mathbb{N}$.
\end{lemma}
\proof{\textup{Proof.}}
To reduce notation, write $d=\dim_{E}\left(\mathcal{F}, T^{-1} \right)$
and $w_{t}=w_{t}(A_{t})$. Reorder the sequence 
$\left(w_{1},...,w_{T}\right)\rightarrow\left(w_{i_{1}},...,w_{i_{T}}\right)$
where $w_{i_{1}}\geq w_{i_{2}}\geq...\geq w_{i_{T}}$. We have 
\[
\sum_{t=1}^{T}w_{\mathcal{F}_t}(A_{t})
=\sum_{t=1}^{T}w_{i_{t}}
=\sum_{t=1}^{T}w_{i_{t}}\mathbf{1}\left\{w_{i_{t}}\leq T^{-1}\right\}
+\sum_{t=1}^{T}w_{i_{t}}\mathbf{1}\left\{w_{i_{t}}> T^{-1}\right\}
\leq 1+\sum_{t=1}^{T}w_{i_{t}}\mathbf{1}\left\{w_{i_{t}}\geq T^{-1}\right\}. 
\]

We know $w_{i_{t}}\leq C$. In addition, $w_{i_{t}}>\epsilon \iff \sum_{k=1}^{T} 
\mathbf{1}\left(w_{\mathcal{F}_k}\left(A_k\right) >\epsilon \right) \geq t$. By Proposition 
\ref{prop: num_times_threshold_exceeded},  this can only occur if $t < \left(\frac{4\beta_T}{\epsilon^2}+1\right)\dim_{E}(\mathcal{F}, \epsilon)$. 
For $\epsilon \geq T^{-1}$, 
$\dim_{E}(\mathcal{F}, \epsilon) \leq \dim_{E}(\mathcal{F}, T^{-1})=d$, since $\dim_{E}\left(\mathcal{F}, \epsilon' \right)$ is nonincreasing in $\epsilon'$. Therefore, when $w_{i_t}>\epsilon \geq T^{-1}$, 
$t<\left(\frac{4\beta_T}{\epsilon^2}+1\right)d $ which implies $\epsilon < \sqrt{\frac{4\beta_{T}d}{t-d}}$.
This shows that if $w_{i_t}>T^{-1}$, then 
$w_{i_t} \leq \min \left\{C,  \sqrt{\frac{4\beta_{T}d}{t-d}} \right\}$.  Therefore, 

\[
\sum_{t=1}^{T}w_{i_{t}}\mathbf{1}\left\{w_{i_{t}}> T^{-1}\right\}
\leq dC+\sum_{t=d+1}^{T}\sqrt{\frac{4d\beta_{T}}{t-d}}\leq dC+2\sqrt{d\beta_{T}}\intop_{t=0}^{T}\frac{1}{\sqrt{t}}dt=dC+
4\sqrt{d \beta_{T}T} \Halmos
\]
\endproof

Our next result, which follows from Lemma \ref{lem: risk under condition 1}, Lemma \ref{lem: sum of widths}, and 
Proposition \ref{prop: least squares bound}, establishes a Bayesian regret bound.
\begin{proposition}
\label{prop: Bayesian regret bound in terms of beta}
For all $T \in \mathbb{N}$, $\alpha > 0$ and $\delta \leq 1/2T$, 
$${\rm BayesRegret}\left(T,\,\pi^{\rm PS}\right) \leq 1+
\left[{\rm dim}_E\left(\mathcal{F}, T^{-1} \right)+1\right]C 
+4\sqrt{{\rm dim}_E\left(\mathcal{F}, T^{-1} \right) \beta^{*}_{T} \left(\mathcal{F}, \alpha, \delta \right) T}.$$
\end{proposition}

Using bounds on $\beta_{t}^{*}$ provided in the previous section together with
Proposition \ref{prop: Bayesian regret bound in terms of beta} yields Bayesian regret bounds that depend on $\mathcal{F}$
only through the eluder dimension and either the cardinality or Kolmogorov dimension.  The following proposition
provides such bounds.
\begin{proposition}
\label{prop: risk bound} 
For any fixed class of functions $\mathcal{F}$,  
$${\rm BayesRegret}\left(T,\,\pi^{\rm PS}\right) \leq
 1+\left[{\rm dim}_E\left(\mathcal{F}, T^{-1} \right)+1\right]C+ 
16\sigma \sqrt{  {\rm dim}_E\left(\mathcal{F},  T^{-1} \right) 
\left(1+o(1)+{\rm dim}_K\left(\mathcal{F} \right) \right)
 \log(T) T
}    $$
for all $T\in \mathbb{N}$. Further, if $\mathcal{F}$ is finite then
$${\rm BayesRegret}\left(T,\,\pi^{\rm PS}\right) \leq 1+\left[{\rm dim}_E\left(\mathcal{F}, T^{-1} \right)+1\right]C
+8\sigma\sqrt{2{\rm dim}_E\left(\mathcal{F}, T^{-1} \right) \log\left(2\left|\mathcal{F}\right|T\right)T},$$
for all $T \in \mathbb{N}$.
\end{proposition}

The next two examples show how the first Bayesian regret bound of Proposition \ref{prop: risk bound} specializes
to $d$-dimensional linear and generalized linear models. For each of these examples, a bound on 
${\rm dim}_E\left(\mathcal{F}, \epsilon \right)$ is provided in the appendix. 

\begin{example}
{\bf Linear Models:}
Consider the case of a $d$-dimensional linear model $f_{\rho}(a):=\left\langle \phi(a),\,\rho\right\rangle$.
Fix $\gamma = \sup_{a \in \mathcal{A}} \| \phi(a)\|$ and $s = \sup_{\rho \in \Theta} \|\rho\|$.  
Then, 
${\rm dim}_E(\mathcal{F}, \epsilon) = O(d\log(1/\epsilon))$ and ${\rm dim}_K(\mathcal{F}) = O(d)$. 
Proposition \ref{prop: risk bound} therefore yields an $O(d\sqrt{T}\log(T))$ Bayesian regret bound.
This is tight to within a factor of $\log T$ \cite{rusmevichientong2010linearly}, and matches 
the best available bound for a linear UCB algorithm \cite{abbasi2011improved}.  
\end{example}

\begin{example}
{\bf Generalized Linear Models:}
Consider the case of a $d$-dimensional generalized linear model $f_{\theta}(a):=g\left(\left\langle \phi(a),\,\theta\right\rangle\right)$ where
$g$ is an increasing Lipschitz continuous function.  
Fix $\gamma = \sup_{a \in \mathcal{A}} \| \phi(a)\|$ and $s = \sup_{\rho \in \Theta} \|\rho\|$.
Then, ${\rm dim}_K(\mathcal{F}) = O(d)$ and
${\rm dim}_E(\mathcal{F}, \epsilon) = O(r^2 d\log(r \epsilon))$,
where $r= \sup_{\tilde{\theta},a}g'(\langle \phi(a),
\,\tilde{\theta}\rangle ) / \inf_{\tilde{\theta},a}g'(\langle \phi(a),\,\tilde{\theta}\rangle)$
bounds the ratio between the maximal and minimal slope of $g$. 
Proposition \ref{prop: risk bound} yields an $O(rd\sqrt{T} \log (rT))$ 
Bayesian regret bound. We know of no other guarantee for posterior sampling when applied to generalized linear models. 
In fact, to our knowledge, this bound is a slight improvement over the strongest Bayesian regret bound available for any
algorithm in this setting. The regret bound of  \citet{filippi2010parametric} translates to an $O(rd\sqrt{T}\log^{3/2}(T))$ 
Bayesian regret bound. 
\end{example}

One advantage of studying posterior sampling in a general framework is 
that it allows bounds to be obtained for specific classes of models by specializing more general results. 
This advantage is highlighted by the ease of developing a performance guarantee for generalized linear models.
The problem is reduced to one of bounding the eluder dimension, and such a bound follows
almost immediately from the analysis of linear models.  In prior literature, 
extending results from linear to generalized linear models required significant technical developments, 
as presented in \citet{filippi2010parametric}.

\subsection{Relation to the Vapnik-Chervonenkis Dimension.}
\label{subsec: vc}

To close our section on general bounds, we discuss important differences between 
our new notion of eluder dimension and complexity measures used in the 
analysis of supervised learning problems.  We begin with an example
that illustrates how a class of functions that is learnable in constant time in 
a supervised learning context may require an arbitrarily long duration when learning to optimize.
\begin{example}
Consider a finite class of binary-valued functions $\mathcal{F}=\left\{f_\rho:\mathcal{A} \mapsto \{0,1\}\ |\ \rho \in \{1,\ldots,n\}\right\}$
over a finite action set $\mathcal{A} = \{1,\ldots,n\}$.  Let $f_\rho(a) = {\bf 1}(\rho = a)$, so that each function
is an indicator for an action.  To keep things simple, assume that $R_t = f_\theta(A_t)$, so that there is no noise.
If $\theta$ is uniformly distributed over $\{1,\ldots, n\}$, it is easy to see that the Bayesian regret of posterior sampling  
grows linearly with $n$.  For large $n$, until $\theta$ is discovered, each sampled action is unlikely to reveal 
much about $\theta$ and learning therefore takes very long.

Consider the closely related supervised learning problem in which at each time step an action $\tilde{A}_t$
is sampled uniformly from $\mathcal{A}$ and the mean--reward value $f_\theta(\tilde{A}_t)$ is observed.
For large $n$, the time it takes to effectively learn to predict $f_\theta(\tilde{A}_t)$ given
$\tilde{A}_t$ does not depend on $t$.  In particular, prediction error converges to $1/n$ in constant time.  
Note that predicting $0$ at every time already achieves this low level of error.
\end{example}

In the preceding example, the $\epsilon$-eluder dimension is $n$ for $\epsilon \in (0,1)$.  On the
other hand, the Vapnik-Chervonenkis (VC) dimension, which characterizes the 
sample complexity of supervised learning, is $1$.
To highlight conceptual differences between the eluder dimension and the VC dimension, we 
will now define VC dimension in a way analogous to how we defined
eluder dimension.  We begin with a notion of independence.
\begin{definition}
An action $a$ is {\it VC-independent} of $\tilde{\mathcal{A}}\subseteq\mathcal{A}$
if for any $f,\,\tilde{f}\in\mathcal{F}$ there exists some $\bar{f}\in\mathcal{F}$
which agrees with $f$ on $a$ and with $\tilde{f}$ on $\tilde{\mathcal{A}}$; that is, 
$\bar{f}(a)=f(a)$ and $\bar{f}(\tilde{a})=\tilde{f}(\tilde{a})$ for all $\tilde{a}\in\tilde{\mathcal{A}}$.
Otherwise, $a$ is {\it VC-dependent} on $\tilde{\mathcal{A}}$.
\end{definition}
By this definition, an action $a$ is said to be VC-dependent on $\tilde{\mathcal{A}}$
if knowing the values $f\in\mathcal{F}$ takes on $\tilde{\mathcal{A}}$
\textit{could restrict} the set of possible values at $a$. This notion of
independence is intimately related to the VC dimension of a class
of functions. In fact, it can be used to define VC dimension.
\begin{definition}
The VC dimension of a class of binary-valued functions
with domain $\mathcal{A}$ is the largest cardinality of a set $\tilde{\mathcal{A}}\subseteq\mathcal{A}$
such that every $a\in\tilde{\mathcal{A}}$ is VC-independent of $\tilde{\mathcal{A}}\backslash\left\{ a\right\} $. 
\end{definition}
In the above example, any two actions are VC-dependent because knowing
the label of one action could completely determine the value of the
other action. However, this only happens if the sampled action has
label 1. If it has label 0, one cannot infer anything about the value
of the other action. Instead of capturing the fact that one \textit{could}
gain useful information about the reward function through exploration, we need a 
stronger requirement that guarantees one \textit{will} gain useful information 
through exploration.  Such a requirement is captured by the following concept.
\begin{definition}
An action $a$ is {\it strongly-dependent} on a set of actions $\tilde{\mathcal{A}} \subseteq \mathcal{A}$
if any two functions $f,\,\tilde{f}\in\mathcal{F}$ that agree on
$\tilde{\mathcal{A}}$ agree on $a$; that is, the set 
$\{ f(a):\, f(\tilde{a})=\tilde{f}(\tilde{a})\,\,\,\forall\tilde{a}\in\tilde{\mathcal{A}}\} $
is a singleton. An action $a$ is {\it weakly independent} of $\tilde{\mathcal{A}}$
if it is not strongly-dependent on $\mathcal{\tilde{A}}$. 
\end{definition}
According to this definition, $a$ is strongly dependent on $\tilde{\mathcal{A}}$
if knowing the values of $f$ on $\tilde{\mathcal{A}}$ completely
determines the value of $f$ on $a$. While the above definition is
conceptually useful, for our purposes it is important to capture approximate
dependence between actions.  Our definition of eluder dimension achieves this goal
by focusing on the possible difference $f(a)-\tilde{f}(a)$ between two functions that 
approximately agree on $\tilde{\mathcal{A}}$.

\section{Simulation Results.}
\label{sec:simulation}

In this section, we compare the performance in simulation of posterior sampling to that of 
UCB algorithms that have been proposed in the recent literature. 
Our results demonstrate that posterior sampling significantly outperforms these 
algorithms. Moreover, we identify a clear cause for the large discrepancy: 
confidence sets proposed in the literature are too loose to attain good performance. 

\begin{figure}[H]
\includegraphics[width=6in]{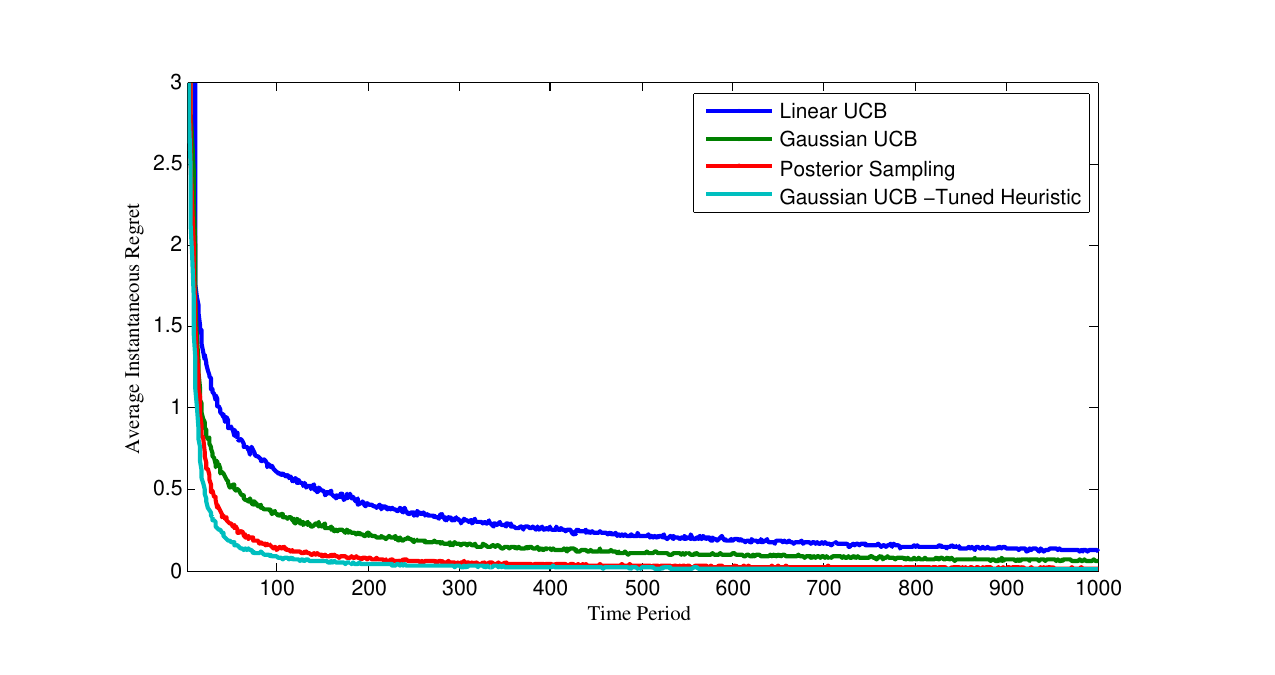}
\caption{Portion of regret attributable to each time period.}\label{fig: average regret}
\end{figure}

We consider the linear model $f_{\theta}(a)=\left\langle \phi(a),\,\theta\right\rangle $
where $\theta\in\mathbb{R}^{10}$ follows a multivariate Gaussian
distribution with mean vector $\mu=0$ and covariance matrix $\Sigma=10I$.
The noise terms $\epsilon_t:=R_t-f_\theta(A_t)$ follow 
a standard Gaussian distribution. There are 100 actions with feature vector components drawn
uniformly at random from $[-1/\sqrt{10}, 1/\sqrt{10}]$, and 
$\mathcal{A}_t=\mathcal{A}$ for each $t$.  Figure \ref{fig: average regret} shows the portion 
$\left\langle \phi(A^{*}_t),\,\theta\right\rangle -\left\langle \phi(A_{t}),\,\theta\right\rangle$ of regret 
attributable to each time period $t$ in the first 1000 time periods.  The results are averaged 
across 5000 trials. 

Several UCB algoriths are suitable for such problems, including those of
\cite{rusmevichientong2010linearly, abbasi2011improved, srinivas2012information}.   
While the confidence bound of \cite{rusmevichientong2010linearly}
is stronger than that of \cite{dani2008stochastic},  it is still too loose and the resulting linear UCB 
algorithm hardly improves its performance over the 1000 period time horizon. We display the results
only of the more competitive UCB algorithms.  The line labeled ``linear UCB'' displays the results
of the algorithm proposed in \cite{abbasi2011improved}, which incurred average regret of 339.7. 
The algorithm of \cite{srinivas2012information} is labeled ``Gaussian UCB,'' and incurred average regret 198.7. 
Posterior sampling, on the other hand, incurred average regret of only 97.5. 

Each of these UCB algorithms uses a confidence bound that was derived through stochastic analysis. 
The Gaussian linear model has a clear structure, however, which suggests upper confidence 
bounds should take the form $U_t(a)=\mu_{t-1}(a)+\sqrt{\beta}\sigma_{t-1}(a)$ where $\mu_{t-1}(a)$
and $\sigma_{t-1}(a)$ are the posterior mean and standard deviation at $a$. The final algorithm we consider
ignores theoretical considerations, and tunes the parameter $\beta$ to minimize the average regret 
over the first 1000 periods. 
The average regret of the algorithm was only 68.9, a dramatic improvement over
\cite{abbasi2011improved}, and \cite{srinivas2012information}, and even outperforming posterior sampling. 
On the plot shown below, these results are labeled ``Gaussian UCB - Tuned Heuristic.'' Note such 
tuning requires the time horizon to be fixed and known. 

In this setting, the problem of choosing upper-confidence bounds reduces to choosing a single confidence 
parameter $\beta$.  For more complicated problems, however,
significant analysis may be required  to choose a structural form for confidence sets. The results in this section suggest
that it can be quite challenging to use such analysis to derive  confidence bounds that lead to strong empirical performance.
In particular, this is challenging even for linear models.  For example, the paper \cite{abbasi2011improved} uses sophisticated tools from the 
study of multivariate self-normalized martingales to derive a confidence bound that is stronger than 
those of \cite{dani2008stochastic} or \cite{rusmevichientong2010linearly}, but their algorithm still incurs 
about three and a half times the regret of posterior sampling. This highlights a crucial advantage
of posterior sampling that we have emphasized throughout this paper; it effectively 
separates confidence bound {\it analysis} from algorithm {\it design}.

Finally, it should be noted that the algorithms of \cite{abbasi2011improved, srinivas2012information} have free parameters that 
must be chosen by the user.  We have attempted to set these values in a way that minimizes average regret over the 1000 period
time horizon. Both algorithms construct confidence bounds that hold with a pre-specified probability $1-\delta\in [0,1] $. Higher levels
of $\delta$ lead to lower upper-confidence bounds, which we find improves performance. We set $\delta=1$ to minimize
the average regret of the algorithms. The algorithm of \cite{abbasi2011improved} requires two other choices. We used a line search to
set the algorithm's regularization parameter to the level $\lambda=.025$, which minimizes cumulative regret. 
The algorithm of  \cite{abbasi2011improved} also requires a uniform upper bound on 
$\left\Vert \theta \right\Vert$, but the Gaussian distribution is unbounded.
We avoid this issue by providing the actual realized value $\left\Vert \theta \right\Vert$ as an input to algorithm.

\section{Conclusion.}

This paper has considered the use of a simple posterior sampling algorithm for learning to optimize actions when the 
decision maker is uncertain about how his actions influence performance.
We believe that, particularly for difficult problem instances,
this algorithm offers significant potential advantages because of its design simplicity and computational tractability. 
Despite its great potential, not much is known about posterior sampling when there are dependencies between actions. Our work has taken a significant step toward remedying this gap. We showed that the Bayesian regret of posterior sampling can be decomposed in terms of confidence sets, which allowed us to establish a number of new results on posterior sampling by leveraging prior work on UCB algorithms. We then used this regret decomposition to analyze posterior sampling in a very general framework, and developed Bayesian regret bounds that depend on a new notion of dimension. 

In constructing these bounds, we have identified two factors that control the hardness of a particular multi-armed bandit problem. First, an agent's ability to quickly attain near-optimal performance depends on the extent to which the reward value at one action can be inferred by sampling other actions. However, in order to select an action the agent must make inferences about many possible actions, and an error in its evaluation of any one could result in large regret. Our second measure of complexity controls for the difficulty of maintaining appropriate confidence sets simultaneously at every action. While our bounds are nearly tight in some cases, further analysis is likely to yield stronger 
results in other cases.  We hope, however, that our work provides a conceptual foundation for the study of such problems, and inspires further investigation.

%
%
%
 \begin{APPENDICES}
\section{Details Regarding Lemma \ref{lem: finite case confidence}.} \label{sec: simpleconcentration}
Lemma \ref{lem: finite case confidence} follows as a special case of Theorem 1 of \citet{abbasi2011improved}, which is much more general. Note that because reward noise $R_{t}-f_{\theta}(A_t)$ is bounded in [-1,1], it is 1-subgaussian. Equation 12 in \citet{abbasi2011improved} gives a specialization of Theorem 1 to the problem we consider. It states that for any $\delta>0$, with probability at least $1-\delta$, 

$$\left| \sum_{k=1}^{t} \left( \mathbf{1}_{\{A_k=a\}} (R_{k}-f_{\theta}(a)  \right) \right|  \leq \sqrt{\left(1+N_t(a)\right)\left(1+2\log\left(\frac{(1+N_{t}(a))^{1/2}}{\delta}\right) \right)}  \,\,\,\,\,\,\, \forall t\in \mathbb{N}.$$
We choose $\delta=1/T$ and use that for $t\leq T$,  $N_{t}(a)\leq T-1$ to show that with probability at least $1/T$, 
$$\left| \sum_{k=1}^{t} \left( \mathbf{1}_{\{A_k=a\}} (R_{k}-f_{\theta}(a)  \right) \right|  \leq \sqrt{\left(1+N_t(a)\right)\left(1+3\log(T) \right)}  \,\,\,\,\,\,\, \forall t\in \{1,..,T\}.$$
Since $1+N_t(a) \leq 2N_t(a)$ whenever $a$ has been played at least once, with probability at least $1/T$,
$$\left| \sum_{k=1}^{t} \left( \mathbf{1}_{\{A_k=a\}} (R_{k}-f_{\theta}(a)  \right) \right|  \leq \sqrt{\left(N_t(a)\right)\left(2+6\log(T) \right)}  \,\,\,\,\,\,\, \forall t\in \{1,..,T\}.$$

\section{Proof of Confidence bound.}
\subsection{Preliminaries: Martingale Exponential Inequalities.}

Consider random variables $\left(Z_{n} | n\in\mathbb{N}\right)$ adapted
to the filtration $\left(\mathcal{H}_{n}:\, n=0,1,...\right)$. Assume
$\mathbb{E}\left[\exp\left\{ \lambda Z_{i}\right\} \right]$ is finite
for all $\lambda$. Define the conditional mean 
$\mu_{i}=\mathbb{E}\left[Z_{i}\mid\mathcal{H}_{i-1}\right]$.
We define the conditional cumulant generating function of the centered
random variable $\left[Z_{i}-\mu_{i}\right]$ by 
$\psi_{i}\left(\lambda\right)=\log\mathbb{E}\left[\exp\left(\lambda\left[Z_{i}-\mu_{i}\right]\right)\mid\mathcal{H}_{i-1}\right]$.
Let 
\[
M_{n}(\lambda)=\exp\left\{ \sum_{i=1}^{n}\lambda\left[Z_{i}-\mu_{i}\right]-\psi_{i}\left(\lambda\right)\right\}.
\]

\begin{lemma}
$\left( M_{n}(\lambda) \vert n\in\mathbb{N} \right)$ is a Martinagale, and $\mathbb{E}M_{n}(\lambda)=1$. \end{lemma}
\proof{\textup{Proof.}}
By definition, $$\mathbb{E}[M_{1}(\lambda) \vert \mathcal{H}_0 ]=
\mathbb{E}[\exp\left\{ \lambda\left[Z_{1}-\mu_{1}\right]-\psi_{1}\left(\lambda\right)\right\} \vert \mathcal{H}_0] =\mathbb{E}[\exp\left\{ \lambda\left[Z_{1}-\mu_{1}\right]\right\} \vert \mathcal{H}_0]/\exp\left\{ \psi_{1}\left(\lambda\right)\right\} =1.$$
Then, for any $n\geq2$ 
\begin{eqnarray*}
\mathbb{E}\left[M_{n}(\lambda)\mid\mathcal{H}_{n-1}\right] & = & \mathbb{E}\left[\exp\left\{ \sum_{i=1}^{n-1}\lambda\left[Z_{i}-\mu_{i}\right]-\psi_{i}\left(\lambda\right)\right\} \exp\left\{ \lambda\left[Z_{n}-\mu_{n}\right]-\psi_{n}\left(\lambda\right)\right\} \mid\mathcal{H}_{n-1}\right]\\
& = & \exp\left\{ \sum_{i=1}^{n-1}\lambda\left[Z_{i}-\mu_{i}\right]-\psi_{i}\left(\lambda\right)\right\} \mathbb{E}\left[\exp\left\{ \lambda\left[Z_{n}-\mu_{n}\right]-\psi_{n}\left(\lambda\right)\right\} \mid\mathcal{H}_{n-1}\right]\\
& = & \exp\left\{ \sum_{i=1}^{n-1}\lambda\left[Z_{i}-\mu_{i}\right]-\psi_{i}\left(\lambda\right)\right\}=M_{n-1}(\lambda). \Halmos \\
\end{eqnarray*}
\endproof

\begin{lemma}
\label{lem: exponential inequality}
For all $x\geq0$ and $\lambda\geq0$, $\mathbb{P}\left(\sum_{1}^{n}\lambda Z_{i}\leq x+\sum_{1}^{n}\left[\lambda\mu_{i}+\psi_{i}\left(\lambda\right)\right]\,\,\,\,\mbox{\ensuremath{\forall}}n\in\mathbb{N}\right)\geq1-e^{-x}$. \end{lemma}
\proof{\textup{Proof.}}
For any $\lambda$, $M_{n}(\lambda)$ is a martingale with $\mathbb{E}M_{n}\left(\lambda\right)=1$.
Therefore, for any stopping time $\tau$, $\mathbb{E}M_{\tau\wedge n}\left(\lambda\right)=1$.
For arbitrary $x\ge0$, define $\tau_{x}=\inf\left\{ n\geq0\mid M_{n}\left(\lambda\right)\geq x\right\} $
and note that $\tau_{x}$ is a stopping time corresponding to the
first time $M_{n}$ crosses the boundary at $x$. Then, $\mathbb{E}M_{\tau_{x}\wedge n}(\lambda)=1$
and by Markov's inequality: 
\[
x\mathbb{P}\left(M_{\tau_{x}\wedge n}\left(\lambda\right)\geq x\right)\leq\mathbb{E}M_{\tau_{x}\wedge n}(\lambda)=1.
\]
We note that the event $\left\{ M_{\tau_{x}\wedge n}\left(\lambda\right)\geq x\right\} =\bigcup_{k=1}^{n}\left\{ M_{k}(\lambda)\geq x\right\} $.
So we have shown that for all $x\geq0$ and $n\geq1$ 
\[
\mathbb{P}\left(\bigcup_{k=1}^{n}\left\{ M_{k}(\lambda)\geq x\right\} \right)\leq\frac{1}{x}.
\]
Taking the limit as $n\rightarrow\infty$, and applying the monotone
convergence theorem shows $\mathbb{P}\left(\bigcup_{k=1}^{\infty}\left\{ M_{k}(\lambda)\geq x\right\} \right)\leq\frac{1}{x}$,
Or, $\mathbb{P}\left(\bigcup_{k=1}^{\infty}\left\{ M_{k}(\lambda)\geq e^{x}\right\} \right)\leq e^{-x}$.
This then shows, using the definition of $M_{k}(\lambda)$, that 
\[
\mathbb{P}\left(\bigcup_{n=1}^{\infty}\left\{ \sum_{i=1}^{n}\lambda\left[Z_{i}-\mu_{i}\right]-\psi_{i}\left(\lambda\right)\geq x\right\} \right)\leq e^{-x}.
\]

\endproof

\subsection{Proof of Lemma \ref{lem: least squares bound}.}
\begin{mytheorem}[Lemma \ref{lem: least squares bound}.]$a=a$.
For any $\delta > 0$ and $f:\mathcal{A} \mapsto \mathbb{R}$, 
with probability at least $1-\delta$,
\[
L_{2,t}(f)\geq L_{2,t}(f_{\theta})+\frac{1}{2}\left\Vert f-f_{\theta}\right\Vert _{2,E_{t}}^{2}-4\sigma^{2}\log\left(1/\delta\right)
\]
simultaneously for all $t \in \mathbb{N}$.
\end{mytheorem}
We will transform our problem in order to apply the general exponential martingale result shown above. 
We set $\mathcal{H}_{t-1}$ to be the $\sigma$-algebra generated by $(H_t, A_t, \theta)$.
By previous assumptions, $\epsilon_{t}:=R_t-f_{\theta}(A_t)$ satisfies $\mathbb{E}[\epsilon_t| \mathcal{H}_{t-1}]=0$
and $\mathbb{E}\left[\exp\left\{ \lambda\epsilon_{t}\right\} \mid{\mathcal{H}}_{t-1}\right]\leq\exp\left\{ \frac{\lambda^{2}\sigma^{2}}{2}\right\} $ a.s. for all $\lambda$. Define $Z_{t}=\left(f_{\theta}\left(A_{t}\right)-R_{t}\right)^{2}-\left(f\left(A_{i}\right)-R_{t}\right)^{2}$
\proof{\textup{Proof.}}
By definition $\sum_{1}^{T}Z_{t} = L_{2,T+1}(f_\theta)-L_{2,T+1}(f)$. Some calculation shows that $Z_{t}=-\left(f(A_{t})-f_{\theta}(A_{t})\right)^{2}+2\left(f\left(A_{t}\right)-f_{\theta}\left(A_{t}\right)\right)\epsilon_{t}$.
Therefore, the conditional mean and conditional cumulant generating function satisfy: 
\begin{eqnarray*}
\mu_{t} & = & \mathbb{E}\left[Z_{t}\mid\mathcal{H}_{t-1}\right]=-\left(f\left(A_{t}\right)-f_{\theta}\left(A_{t}\right)\right)^{2}\\
 \psi_{t}(\lambda) & = & \log\mathbb{E}\left[\exp\left(\lambda\left[Z_{t}-\mu_{t}\right]\right)\mid\mathcal{H}_{t-1}\right] \\
& = & \log\mathbb{E}\left[\exp\left(2\lambda\left(f\left(A_{t}\right)-f_{\theta}\left(A_{t}\right)\right)\epsilon_{t}\right)\mid\mathcal{H}_{t-1}\right]
\leq  \frac{(2\lambda\left[f\left(A_{t}\right)-f_{\theta}\left(A_{t}\right)\right])^{2}\sigma^{2}}{2}
\end{eqnarray*}
Applying Lemma \ref{lem: exponential inequality} shows that for all $x\geq0$,
$\lambda\geq0$
\[
\mathbb{P}\left(\sum_{k=1}^{t}\lambda Z_{k}\leq x-\lambda\sum_{k=1}^{t}\left(f\left(A_{k}\right)-f_{\theta}\left(A_{k}\right)\right)^{2}+\frac{\lambda^{2}}{2}\left(2f\left(A_{k}\right)-2f_{\theta}\left(A_{k}\right)\right)^{2}\sigma^{2}\,\,\,\,\mbox{\ensuremath{\forall}}t\in\mathbb{N}\right)\geq1-e^{-x}.
\]
 Or, rearranging terms 
\[
\mathbb{P}\left(\sum_{k=1}^{t}Z_{k}\leq\frac{x}{\lambda}+\sum_{k=1}^{t}\left(f\left(A_{k}\right)-f_{\theta}\left(A_{k}\right)\right)^{2}\left(2\lambda\sigma^{2}-1\right)\,\,\,\,\mbox{\ensuremath{\forall}}t\in\mathbb{N}\right)\geq1-e^{-x}.
\]
 Choosing $\lambda=\frac{1}{4\sigma^{2}}$, $x=\log\frac{1}{\delta}$,
and using the definition of $\sum_{1}^{t}Z_{k}$ implies 
\[
\mathbb{P}\left(L_{2,t}(f)\geq L_{2,t}(f_{\theta})+\frac{1}{2}\left\Vert f-f_{\theta}\right\Vert _{2,E_{t}}^{2}-4\sigma^{2}\log\left(1/\delta\right)\,\,\,\,\mbox{\ensuremath{\forall}}t\in\mathbb{N}\right)\geq1-\delta.
\]
\endproof

\subsection{Least Squares Bound - Proof of Proposition \ref{prop: least squares bound}.}
\begin{mytheorem}[Proposition \ref{prop: least squares bound}.]
For all $\delta > 0$ and $\alpha > 0$, if 
$\mathcal{F}_t=\left\{ f\in\mathcal{F}:\,\left\Vert f-\hat{f}_t^{LS}\right\Vert _{2,E_t}
\leq\sqrt{\beta^{*}_t \left( \mathcal{F}, \delta, \alpha\right)}    \right\}$
for all $t \in \mathbb{N}$, then
\[
\mathbb{P}\left(f_{\theta}\in\bigcap_{t=1}^{\infty}\mathcal{F}_{t}\right)\geq1-2\delta.
\]
\end{mytheorem}
\proof{\textup{Proof.}}
Let $\mathcal{F}^{\alpha}\subset\mathcal{F}$ be an $\alpha$--cover
of $\mathcal{F}$ in the sup-norm in the sense that for any $f\in\mathcal{F}$
there is an $f^{\alpha}\in\mathcal{F}^{\alpha}$ such that $\left\Vert f^{\alpha}-f\right\Vert _{\infty}\leq\epsilon$.
By a union bound, with probability at least $1-\delta$, 
\[
L_{2,t}(f^{\alpha})-L_{2,t}(f_{\theta})\geq\frac{1}{2}\left\Vert f^{\alpha}-f_{\theta}\right\Vert _{2,E_{t}}-4\sigma^{2}\log\left(\left|\mathcal{F}^{\alpha}\right|/\delta\right)\,\,\,\,\mbox{\ensuremath{\forall}}t\in\mathbb{N},\, f\in\mathcal{F}^{\alpha}.
\]
Therefore, with probability at least $1-\delta$, for all $t\in\mathbb{N}$ and $f\in\mathcal{F}$:

\begin{eqnarray*}
L_{2,t}(f)-L_{2,t}(f_{\theta})&\geq& \frac{1}{2}\left\Vert f-f_{\theta}\right\Vert _{2,E_{t}}^{2}-4\sigma^{2}\log\left(\left|\mathcal{F}^{\alpha}\right|/\delta\right)\\ 
&+&\underbrace{\min_{f^{\alpha}\in\mathcal{F}^{\alpha}}\left\{ \frac{1}{2}\left\Vert f^{\alpha}-f_{\theta}\right\Vert _{2,E_{t}}^{2}-\frac{1}{2}\left\Vert f-f_{\theta}\right\Vert _{2,E_{t}}^{2}+L_{2,t}(f)-L_{2,t}(f^{\alpha})\right\} }_{\text{Discretization\,\ Error}}.
\end{eqnarray*}
Lemma \ref{lem:discretization}, which we establish in the next section, asserts that with probability at least $1-\delta$
the discretization error is bounded for all $t$ by $\alpha\eta_{t}$
where $\eta_{t}:=t\left[8C+\sqrt{8\sigma^{2}\ln(4t^{2}/\delta)}\right]$.
Since the least squares estimate $\hat{f}_{t}^{LS}$ has lower squared
error than $f_{\theta}$ by definition, we find with probability at
least $1-2\delta$ 
\[
\frac{1}{2}\left\Vert \hat{f}_t^{\rm LS}-f_{\theta}\right\Vert _{2,E_{t}}^{2}\leq4\sigma^{2}\log\left(\left|\mathcal{F}^{\alpha}\right|/\delta\right)+\alpha\eta_{t}.
\]
Taking the infimum over the size of $\alpha$ covers implies: 
\[
\left\Vert \hat{f}_{t}^{LS}-f_{\theta}\right\Vert _{2,E_{t}}\leq\sqrt{8\sigma^{2}\log\left(N(\mathcal{F},\,\alpha,\,\left\Vert \cdot\right\Vert _{\infty})/\delta\right)+2\alpha\eta_{t}} \overset{{\rm def}}{=} \sqrt{\beta^{*}_{t}( \mathcal{F}, \delta, \alpha)}. \Halmos
\]
\endproof

\subsection{Discretization Error.}
\begin{lemma}
\label{lem:discretization}
If $f^{\alpha}$ satisfies $\left\Vert f-f^{\alpha}\right\Vert _{\infty}\leq\alpha$, then with probability
at least $1-\delta$, 
\begin{equation} \label{eq: discretization error}
\left\vert \frac{1}{2}\left\Vert f^{\alpha}-f_{\theta}\right\Vert _{2,E_{t}}^{2}-\frac{1}{2}\left\Vert f-f_{\theta}\right\Vert _{2,E_{t}}^{2}+L_{2,t}(f)-L_{2,t}(f^{\alpha})\right\vert \leq  
\alpha t\left[8C+\sqrt{8\sigma^{2}\ln(4t^{2}/\delta)}\right]  \,\,\ \forall t\in\mathbb{N}.
\end{equation}
\end{lemma}
\proof{\textup{Proof.}}
Since any two functions in $f,f^{\alpha} \in \mathcal{F}$ satisfy $\left\Vert f-f^{\alpha}\right\Vert _{\infty}\leq C$, it is enough to consider
$\alpha \leq C$. We find
$$
\left\vert \left(f^{\alpha}\right)^{2}(a)-\left(f\right)^{2}(a)\right\vert  \leq  \max_{-\alpha\leq y\leq\alpha}
\left\vert \left(f(a)+y\right)^{2}-f(a)^{2} \right\vert =2f(a)\alpha+\alpha^{2}\leq2C\alpha+\alpha^{2}
$$
which implies
\begin{eqnarray*}
\left\vert \left(f^{\alpha}(a)-f_{\theta}(a)\right)^{2}-\left(f(a)-f_{\theta}(a)\right)^{2} \right\vert & = & \left\vert  \left[\left(f^{\alpha}\right)(a)^{2}-f(a)^{2}\right]+2f_{\theta}(a)\left(f(a)-f^{\alpha}(a)\right) \right\vert  \leq 4C\alpha+\alpha^{2} \\
\left\vert \left(R_{t}-f(a)\right)^{2}-\left(R_{t}-f^{\alpha}(a)\right)^{2} \right\vert & = & \left\vert 2R_{t}\left(f^{\alpha}(a)-f(a)\right)+f(a)^{2}-f^{\alpha}(a)^{2} \right\vert   \leq2\alpha\left|R_{t}\right|+2C\alpha+\alpha^{2}\\
\end{eqnarray*}
Summing over $t$, we find that the left hand side of (\ref{eq: discretization error}) is bounded by 
\[
\sum_{k=1}^{t-1}\left(\frac{1}{2}\left[4C\alpha+\alpha^{2}\right]+\left[2\alpha\left|R_{k}\right|+2C\alpha+\alpha^{2}\right]\right)\leq\alpha\sum_{k=1}^{t-1}\left(6C+2\left|R_{k}\right|\right).
\]
Because $\epsilon_{k}$ is sub-Gaussian, $\mathbb{P}\left(\left|\epsilon_{k}\right|>\sqrt{2\sigma^{2}\ln(2/\delta)}\right)\leq\delta$.
By a union bound, $$\mathbb{P}\left(\exists k\, s.t.\,\,\left|\epsilon_{k}\right|>\sqrt{2\sigma^{2}\ln(4t^{2}/\delta)}\right)\leq\frac{\delta}{2}\sum_{1}^{\infty}\frac{1}{k^{2}}\leq\delta.$$
Since $\left|R_{k}\right|\leq C+\left|\epsilon_{k}\right|$ this shows
that with probability at least $1-\delta$ the discretization error
is bounded for all $t$ by $\alpha\eta_{t}$ where 
$\eta_{t}:=t\left[8C+2\sqrt{2\sigma^{2}\ln(4t^{2}/\delta)}\right]$. \Halmos
\endproof

\section{Bounds on Eluder Dimension for Common Function Classes.} 

Definition \ref{def: dimension}, which defines the eluder dimension of a class of functions, can be equivalently written as follows. The $\epsilon$-eluder dimension of a class of functions $\mathcal{F}$ 
is the length of the longest sequence  $a_1,..,a_\tau$ such that for some 
$\epsilon' \geq \epsilon$
\begin{equation}
\label{eq: abbreviated width notation}
w_k:=\sup \left\{  \left(f_{\rho_1}-f_{\rho_1} \right)(a_k) : \sqrt{\sum_{i=1}^{k-1}\left(f_{\rho_1}-f_{\rho_2}\right)^{2}\left(a_{i}\right)}\leq\epsilon' \,\,  \rho_1, \rho_2 \in\Theta\right\} > \epsilon'
\end{equation}
for each $k\leq\tau$.

\subsection{Finite Action Spaces.}
Any action is $\epsilon'$--dependent on itself since 
$$\sup \left\{  \left(f_{\rho_1}-f_{\rho_1} \right)(a) : \sqrt{\left(f_{\rho_1}-f_{\rho_2}\right)^{2}\left(a \right)}\leq\epsilon' \,\,  \rho_1, \rho_2 \in\Theta\right\}\leq\epsilon' .$$ 
Therefore, for all $\epsilon>0$, the $\epsilon$-eluder dimension of $\mathcal{A}$ is bounded by $|\mathcal{A}|$.  

\subsection{Linear Case.}

\begin{proposition} 
\label{prop: linear eluder dimension}
Suppose $\Theta \subset \mathbb{R}^d$ and $f_\theta(a)=\theta^T \phi(a)$. Assume there exist constants $\gamma$, and $S$, such that for all $a\in \mathcal{A}$ and $\rho \in \Theta$, $\left\Vert \rho \right\Vert_2 \leq S$, and $\left\Vert \phi(a) \right\Vert_2 \leq \gamma$. Then ${\rm dim}_E( \mathcal{F}, \epsilon) \leq 
3d \frac{e}{e-1} \ln\left\{3+3 \left(\frac{2S}{\epsilon} \right)^2 \right\}+1$.  
\end{proposition}

To simplify the notation, define $w_k$ as in (\ref{eq: abbreviated width notation}), 
$\phi_k=\phi\left( a_k \right)$, $\rho=\rho_1-\rho_2$, and 
$\Phi_k=\sum_{i=1}^{k-1}\phi_{i} \phi_{i}^{T}$. In this case, 
$\sum_{i=1}^{k-1}\left(f_{\rho_1}-f_{\rho_2}\right)^{2}\left(a_{i}\right)=\rho^{T}\Phi_k \rho$, and by the triangle inequality
 $\left\Vert \rho \right\Vert_2 \leq 2S$. The proof follows by bounding the number of times $w_k > \epsilon'$ can occur. 

\textbf{Step 1:} If $w_k\geq\epsilon'$ then $\phi_k^{T}V_k^{-1}\phi_k\geq\frac{1}{2}$ where $V_k:=\Phi_k+\lambda I$ and $\lambda =\left(\frac{\epsilon'}{2S}\right)^{2} $. 
\proof{\textup{Proof.}}
We find $w_{k}\leq
\max\left\{ \rho^{T}\phi_k:\,\rho^{T}\Phi_{k}\rho\leq (\epsilon')^{2},\,\,\rho^{T}I\rho\leq (2S)^2\right\} 
\leq \max\left\{ \rho^{T}\phi_k :\, \rho^{T}V_{k}\rho_k \leq 2(\epsilon')^{2}   \right\}
= \sqrt{2(\epsilon')^2}\left\Vert  \phi_k \right\Vert_{V_k^{-1}}$.
The second inequality follows because any $\rho$ that is feasible for the first maximization problem
must satisfy $\rho^T V_k \rho \leq (\epsilon')^2 +\lambda (2S)^2=2(\epsilon')^2$. 
By this result, $w_k\geq \epsilon'$ implies $\left\Vert  \phi_k \right\Vert_{V_k^{-1}}^2 \geq 1/2$. \Halmos
\endproof

\textbf{Step 2:} If $w_i \geq \epsilon'$ for each $i<k$ then $\det V_k\geq \lambda^{d}\left(\frac{3}{2}\right)^{k-1}$ and
$\det V_k \leq  \left(\frac{\gamma^{2}\left(k-1\right)}{d}+\lambda \right)^{d}$. 
\proof{\textup{Proof.}}
Since $V_{k}=V_{k-1}+\phi_{k}\phi_{k}^{T}$, using the
Matrix Determinant Lemma,
\[
\det V_{k} =\det V_{k-1}\left(1+\phi_{t}^{T} V_k^{-1}\phi_{t}\right)
\geq \det V_{k-1}\left(\frac{3}{2}\right)
\geq...\geq 
\det\left[\lambda I\right] \left( \frac{3}{2} \right)^{k-1}
= \lambda^d \left( \frac{3}{2} \right)^{k-1}.
\] 
Recall that $\det V_k$ is the product
of the eigenvalues of $V_k$, whereas $\text{trace}\left[ V_k \right]$ is the sum. 
As noted in \cite{dani2008stochastic}, $\det V_k$ is maximized when all eigenvalues are equal.
This implies:
$\det V_k \leq \left(\frac{\text{trace}\left[ V_k \right]}{d}\right)^{d} 
\leq\left(\frac{\gamma^2\left(t-1\right)}{d}+\lambda \right)^{d}$. \Halmos
\endproof

\textbf{Step 3:} Complete Proof.
\proof{\textup{Proof.}}
Manipulating the result of Step 2 shows $k$ must satisfy the inequality: 
$\left(\frac{3}{2}\right)^{\frac{k-1}{d}}\leq \alpha_0\left[\frac{k-1}{d}\right]+1$ where 
$\alpha_0=\left(\frac{ \gamma^2}{\lambda}\right)=\left(\frac{2S \gamma }{\epsilon'}\right)^2$. 
Let $B(x, \alpha)=\max\left\{ B:\,\left(1+x\right)^{B}\leq \alpha B+1\right\} $.
The number of times $w_k > \epsilon'$ can occur is bounded by $d B(1/2, \alpha_0)+1$. 

We now derive an explicit bound on $B(x, \alpha)$ for any $x\leq1$. Note that any $B\geq1$ must satisfy
the inequality: $\ln \left\{1+x\right\}B \leq \ln \left\{1+\alpha \right\} + \ln B$. 
Since $\ln \left\{1+x\right\}  \geq x/(1+x)$, using
the transformation of variables $y=B \left[x/(1+x) \right]$ gives: 
$$ y \leq \ln\left\{1+\alpha \right\}+\ln \frac{1+x}{x}+\ln y \leq 
\ln\left\{1+\alpha \right\}+\ln \frac{1+x}{x}+\frac{y}{e} 
\implies y \leq \frac{e}{e-1}\left(\ln\left\{1+\alpha \right\}+\ln \frac{1+x}{x} \right).$$
This implies $B(x, \alpha) \leq \frac{1+x}{x} \frac{e}{e-1}\left(\ln\left\{1+\alpha \right\}+\ln \frac{1+x}{x} \right)$. 
The claim follows by plugging in $\alpha=\alpha_0$ and $x=1/2$. \Halmos


%

\endproof

\subsection{Generalized Linear Models.}
\begin{proposition}
Suppose $\Theta \subset \mathbb{R}^d$ and $f_\theta(a)=g( \theta^T \phi(a))$ where $g(\cdot)$ 
is a differentiable and strictly increasing function. Assume there exist constants $\underline{h}$,
$\overline{h}$, $\gamma$, and $S$, such that for all $a\in \mathcal{A}$ and $\rho \in \Theta$,
$0< \underline{h} \leq g'( \rho^T \phi(a))\leq \overline{h}$, $\left\Vert \rho \right\Vert_2 \leq S$, 
and $\left\Vert \phi(a) \right\Vert_2 \leq \gamma$.  Then ${\rm dim}_E( \mathcal{F}, \epsilon) \leq 3dr^2 \frac{e}{e-1} \ln\left\{3r^2+3r^2 \left(\frac{2S \overline{h}}{\epsilon} \right)^2 \right\}+1$. 
\end{proposition}

The proof follows three steps which closely mirror those used to prove Proposition \ref{prop: linear eluder dimension}. 

\textbf{Step 1:} If $w_k\geq\epsilon'$ then $\phi_k^{T}V_k^{-1}\phi_k\geq\frac{1}{2r^2}$ 
where $V_k:=\Phi_k+\lambda I$ and $\lambda =\left(\frac{\epsilon'}{2S \underline{h}}\right)^{2} $. 
\proof{\textup{Proof.}}
By definition $w_{k}\leq
\max\left\{g\left( \rho^{T}\phi_k \right)  :\, \sum_{i=1}^{k-1} g\left( \rho^{T} \phi(a_i)  \right)^2 
\leq (\epsilon')^{2},\,\,\rho^{T}I\rho\leq (2S)^2\right\}$. By the uniform bound on $g'(\cdot)$ this is less than
$\max\left\{\overline{h} \rho^{T}\phi_k :\, 
\underline{h}^2 \rho^{T}\Phi_{k}\rho \leq (\epsilon')^{2}, \rho^{T}I\rho\leq (2S)^2  \right\}
\leq \max\left\{\overline{h} \rho^{T}\phi_k :\, \underline{h}^2 \rho^{T}V_{k}\rho \leq 2 (\epsilon')^{2}   \right\}
= \sqrt{2(\epsilon')^2 / r^2}\left\Vert  \phi_k \right\Vert_{V_k^{-1}}$.\Halmos
\endproof

\textbf{Step 2:} If $w_i \geq \epsilon'$ for each $i<k$ then $\det V_k\geq \lambda^{d}\left(\frac{3}{2}\right)^{k-1}$ and
$\det V_k \leq  \left(\frac{\gamma^{2}\left(k-1\right)}{d}+\lambda \right)^{d}$. 

\textbf{Step 3:} Complete Proof.
\proof{\textup{Proof.}}
The above inequalities imply $k$ must satisfy: $\left(1+\frac{1}{2r^2}\right)^{\frac{k-1}{d}} 
\leq \alpha_0 \left[\frac{k-1}{d}\right]$ where $\alpha_0 = \gamma^2 / \lambda$. Therefore, as in the linear case,
the number of times $w_k > \epsilon'$ can occur is bounded by $d B(\frac{1}{2r^2}, \alpha_0)+1$. 
Plugging these constants into the earlier bound
$B(x, \alpha) \leq \frac{1+x}{x} \frac{e}{e-1}\left(\ln\left\{1+\alpha \right\}+\ln \frac{1+x}{x} \right)$
and using $1+x\leq 3/2$ yields the result. \Halmos
\endproof

 \end{APPENDICES}

\section*{Daniel Russo is supported by a  Burt and Deedee McMurty Stanford Graduate Fellowship. This work was supported in part by Award CMMI-0968707 from the National Science Foundation.}


\bibliographystyle{ormsv080} 
\bibliography{LearningToOptimize} 


\end{document}